**Intelligent Conversational Bot for Massive Online Open Courses (MOOCs)**

Ser Ling Lim[1], Ong Sing Goh[2]

Faculty of Information and Communication Technology,
Universiti Teknikal Malaysia Melaka, 76100 Melaka, Malaysia
[2]goh@utem.edu.my, [1]limserling@hotmail.com

**Abstract:** Massive Online Open Courses (MOOCs) which were introduced in 2008 has since drawn attention around the world for both its advantages as well as criticism on its drawbacks. One of the issues in MOOCs which is the lack of interactivity with the instructor has brought conversational bot into the picture to fill in this gap. In this study, a prototype of MOOCs conversational bot, MOOC-bot is being developed and integrated into MOOCs website to respond to the learner's enquiries using text or speech input. MOOC-bot is using the popular Artificial Intelligence Markup Language (AIML) to develop its knowledge base, leverage from AIML's capability to deliver appropriate responses and can be quickly adapted to new knowledge domains. The system architecture of MOOC-bot consists of knowledge base along with AIML interpreter, chat interface, MOOCs website and Web Speech API to provide speech recognition and speech synthesis capability. The initial MOOC-bot prototype has the general knowledge from the past Loebner Prize winner - ALICE, course's frequent asked questions, and a course's content offered by Universiti Teknikal Malaysia Melaka (UTeM). The evaluation of MOOC-bot based on the past competition questions from Chatterbox Challenge (CBC) and Loebner Prize has shown that it was able to provide correct answers most of the time during the test and demonstrated the capability to prolong the conversation. The advantages of MOOC-bot such as able to provide 24-hour service that can serve different time zones, able to have knowledge in multiple domains, and can be shared by multiple sites simultaneously have outweighed its existing limitations.

**Keywords:** Conversation Agent (CA), Artificial Intelligence (AI), Massive Online Open Courses (MOOCs), MOOC-bot

# CHAPTER 1

# INTRODUCTION

## 1.1 Background

Online learning has long existed since the rapid growth of Internet penetration. As Internet usage becomes more economical and popular, online learning has found its way becoming more common and well-accepted. In recent years, a diversified form of online learning model called Massive Online Open Courses (MOOCs) has emerged and is drawing considerable attention from around the world (Chatterjee and Nath, 2014). This can also be evidenced by the research publication on MOOCs which has also grown exponentially from 2008 to 2013 (Aparicio et. al., 2014).

MOOCs were first introduced in year 2008 and many universities have adopted it since then. Among the prominent MOOCs providers are Coursera, edX and Udacity (Peco and Luján-Mora, 2013). Andrew Ng and Daphne Koller of Stanford University founded Coursera platform together where Ng taught "Machine Learning" while Koller taught "Probabilistic Graphical Models". EdX is the MOOCs platform founded by Harvard University partnering with Massachusetts Institute of Technology (MIT). Udacity was conceived out of the experiment which saw Sebastian Thrun and Peter Norvig of Stanford University offered "Introduction to Artificial Intelligence" course online. All of them have since extended partnerships with other universities and institutions, and offering more courses.

Unlike conventional online courses, MOOC differs by having pre-recorded video lectures, online quizzes, homework assignments, midterms, final exams and the integration of social networking. The word "massive" also denotes that it can accommodate large number of participants simultaneously across the world for a single course. MOOC's open learning styles offers flexibility with minimal registration requirements. This platform also provides online community interaction in the learning process. Due to all these characteristics, it is not possible to manage and deliver MOOC in the traditional online course's method (Bassi et. al., 2014).

Malaysia which has high number of Internet users can benefit from MOOCs. According to Economic Report 2013/2014 released by the Ministry of Finance Malaysia (2013), the number of Internet users in Malaysia has grown to 18 million as of end of 2012 and is expected to reach 25 million in 2015. MOOCs also fit as a lifelong learning program because it offers the time flexibility to students with families, full-time jobs, or other possible situations to learn at their convenience. The Malaysia Education Blueprint 2015-2025 for Higher Education which was launched recently also enlisted MOOCs as one of the initiatives in spurring Malaysia's higher education system. This blueprint further pointed out that education should not just be sought in one's youth time but people of all ages should also enrich themselves through constant learning. The call for building "Malaysia education brand globally" in the blueprint indicates the great potential of growing MOOCs in Malaysia by local institutions (Ministry of Education Malaysia, 2015).

However, MOOC's delivery is not without its own challenges. For instance, being accessible worldwide would mean the instructors would not be available all the time due to time zone difference. Furthermore, the large number of participants in a course could result in overwhelming student-instructor ratio. As the consequences, students may not able to get tutoring or help from the instructors easily compare to traditional courses. In this case, conversational bot which can simulate human conversation in natural language text could act as a virtual teacher to provide answers to students whenever needed (Ong et. al., 2014).

In this research, MOOC-bot is developed as the conversational bot agent for the role of interacting with users in MOOCs. MOOC-bot will be using Artificial Intelligence Markup Language (AIML) to develop its knowledge base. AIML is a well-known XML derived language which is used in case-based reasoning and textual pattern matching algorithms. AIML was developed by Dr. Richard Wallace and the Alicebot open source community among 1995 and 2000. It was designed to help in the development of ALICE (Artificial Linguistic Internet Computer Entity) (Mikic et. al., 2009). As AIML is open source and can be quickly adapted to new knowledge domains, many conversational bots used AIML to develop their agents that communicate with users in natural language.

MOOC-bot will be the conversational bot for the subject "MITI 5113 Artificial Intelligence" which is offered by Faculty of Information and Communication Technology from Universiti Teknikal Malaysia Melaka (UTeM). UTeM is the first technical university and one of the public universities in Malaysia. Established in year 2000, UTeM was formerly known as Kolej Universiti Teknikal Kebangsaan Malaysia (KUTKM) before being conferred university status. Currently, UTeM operates from three campuses namely the Main Campus, City Campus and the Technology Campus in Malacca with seven faculties and the blend of local and international students.

## 1.2 Statement of the Purpose

The purpose of this research is to develop conversational bot for MOOC as a course conversational bot using AIML. The target course for this research would be "MITI 5113 Artificial Intelligence" offered by UTeM.

## 1.3 Problem Statement

MOOCs are a new online learning model which can accommodate massive number of participants. This can be seen in "Artificial Intelligence" MOOC offered by Stanford University in 2011 where it attracted 160,000 enrolments (Iqbal et. al., 2014). With usually only one instructor delivering a course, students will not be able to get full attention or help from the instructor easily when they face problems. This will end up being frustrating for the students. On the other hand, the instructors will also feel overwhelm to handle so many enquiries from students at one time. Moreover, if the same question was asked by many different students, it is inefficient for the instructor to repeat the answers many times and thus causing additional workload.

In addition, MOOCs are supposed to offer flexibility to students to learn at their own convenience because they may have full time job or any other commitments. At the same time, MOOCs are also offered to participants from worldwide which means they could be living at places with different time zone. Under such circumstances, the students' learning hour may be different from the instructor's working hour. As such, the students are not able to interact or getting feedback from the instructor immediately. This may demotivate the learner, discourage them to ask questions or seek further clarification from the instructor.

Due to all these problems, MOOC-bot could play the role of assistant to the instructor which can respond to course-related enquiries in live chat and as frequently as needed. MOOC-bot which can provide 24-hour services also can solve the flexibility requirement needed in MOOCs. This means that the students can get response at any time needed by them as long as they are connected to the Internet.

## 1.4 Research Questions

Based on the problem statements mentioned in section 1.3, the research questions for this study are identified as follows:

1.  Does MOOC-bot able to synthesize the questions posed by user in text or speech?

2.  Does MOOC-bot able to provide the answer in text and speech?

3.  Does MOOC-bot able to provide the answers appropriately and prolong the conversation?

## 1.5 Research Objective

The research objectives are related to the research questions. Therefore, the research objectives derived are as follows:

1.  To synthesize the questions posed by the users using text or speech input.

2.  To provide answers to questions in text and speech.

3.  To ensure MOOC-bot is able to deliver appropriate answers and prolong the conversation with the users.

**1.6     Research Scope and Limitation**

This research will have the scope and limitation such as follows:

1)     This research will focus on one of the courses offered by UTeM which is "MITI 5113 Artificial Intelligence".

2)     The questions that can be answered must be within MOOC-bot knowledge base.

3)     The question to be answered must be text or speech input in English.

4)     The speech input and speech answers are only supported using Google     Chrome or Apple Safari web browsers.

**1.7     Significant and Research Contribution**

This research is expected to offer a solution to overcome the existing shortcoming of MOOCs through a conversational bot. The conversational bot can be used in different courses because its knowledge can be added and customized. It also has the big advantage of providing unwearying 24-hour services to students which is a great help in MOOCs. In addition, the conversational bot has the convenience of accepting text or speech input and the replies can also be in text and speech.

**1.8     Organization of the Thesis**

This thesis is organized in the following orders:

Chapter 1: Introduction

This chapter introduces the background of MOOCs, the problems faced this topic, the proposed solution and the objectives of this research.

Chapter 2: Literature Review

This chapter reviews the literatures on MOOCs and the proposed solution which in this case is the conversational bot and its related fields.

Chapter 3: Research Methodology

This chapter discusses the methods used in this research to achieve the objectives and the system architecture.

Chapter 4: Implementation

This chapter explains the process of implementing the system architecture and the development of the prototype.

Chapter 5: Evaluation and Result

This chapter explains how the evaluation is done and displays the results of the evaluation.

Chapter 6: Conclusion and Future Work

This chapter concludes the overall study, discusses the strengths and weaknesses, and the recommendation for future work.

# CHAPTER 2

# LITERATURE REVIEW

## 2.1    Introduction

This chapter reviews Massive Open Online Courses (MOOCs), conversational bots and its related topics. Artificial Intelligence (AI) is the field of study which inspires the exploration of many useful applications using a computer or a machine and one of it is conversational bots. Conversational bots are sometimes referred as conversation robots, conversation agents, chat bot, chatter bots or simply 'bots' but they have the same target which is to generate meaningful conversation with human users using natural language. For this purpose, Turing test (TT) is the well-known measurement to assess the performance of a conversation bot and many parties are using TT as the assessment of conversational bots such as the Loebner Prize and Chatterbox Challenge competition though they may be the modified version of the original TT.

There are many types of conversational bots have been explored and developed over the years. They ranged from general conversational bots which do not specialized in any domain to those subject-matter expert bots. The purpose of these conversational bots can also be different; they may be developed for winning competition such as the Loebner Prize or for commercial needs such as attracting or assisting customers. This chapter also outlined the significant works in conversational bot which created impact or contribution to the development in this field. In addition, Artificial Intelligence Markup Language (AIML) and Web Speech Application Programming Interface (API) which are part of MOOC-bot implementation are also further elaborated.

## 2.2    Massive Open Online Courses

MOOCs were started in 2008 through the course, Connectivism and Connective Knowledge (CCK08) which was created and facilitated by George Siemens and Stephen Downes (Fini, 2009; Lowe, 2014). Since then there has been tremendous development in MOOCs with millions of dollars of grants funded the experiments of variety principles, courses, and tools in MOOCs. Until today it is still developing and there is no sign of winding down.

MOOC is considered an education revolution as it differs from traditional course by allowing large enrollments of students from all over the world regardless of prior learning. Some even defined it as courses being offered free of charge to anyone who are interested to learn. However, this is still a debatable topic as there is still a need of a business model to keep MOOCs sustainable (Dellarocas and Van Alstyne, 2013; Valentin, 2014). Anyhow, MOOCs is seen as the route to cheap education which will always be attractive to the public (Lowe, 2014).

As far as the model is concerned, Siemen (2013) has classified three types of MOOCs: xMOOCs, cMOOCs, and quasi-MOOCs. xMOOCs replicate the traditional university model where the teacher plays as the expert and learners become the consumer of the knowledge. Most universities are the driving force behind this model. These include Stanford through Coursera, MIT and Harvard through edX, Udacity, and many elite American institutions. Assignments are computer-graded and direct instructor feedback is scarce in xMOOCs due to the large number of students. cMOOCs are based on connectivism that attempt to group the learners to interact with each other and share their personal knowledge in order to learn together. In contrast with xMOOCs, cMOOCs is not the type of teacher-student knowledge transfer approach but more of knowledge sharing between participants. It emphasizes on autonomy on choosing what content or skills to be learnt by the participant and the openness in the activities that participant can pursue using numerous tools and technologies. CCK08 was the MOOC delivered in cMOOC format. Quasi-MOOCs are more recognized as web-based tutorials in Open Educational Resources (OER) such as those offered by the Khan Academy and MIT's OpenCourseWare (OCW). This model which intended for OER has neither automated grading such as those in xMOOCs nor social interaction connectivity of cMOOCs, but more of a range of loosely related educational resources collected together.

MOOCs have various advantages and challenges. Jurenas (2014) has described the increased accessibility, enhanced student engagement, and lifelong learning opportunity as the advantages of MOOCs. LeCounte and Johnson (2014) opined that MOOCs provides the flexibility for working employees to train on specific skill-sets and also address skills gap problem for companies which need qualified applicants in fields such as accounting and finance, engineering, IT, etc. On the other hand, the issues that MOOCs faced are the challenges of high dropout rates, sustainability, cheating and plagiarism (Siemen, 2013). On top of these, Jurenas (2014) also included the lack of individual instruction and impact on professional life of the instructor as the additional challenges. LeCounte and Johnson (2014) were of the view that the creditability of MOOCs is questionable due to the industry standards for pedagogy and evaluation criteria, not to mention the credit transferability of MOOCs among institutions.

There are various arguments regarding the adoption of MOOCs with some supporting it and some against it as well as those suggest for improvement. Samuels (2014) is one of those against MOOCs with the views that learners tend to multitask and not concentrate in self-paced learning which takes them twice as long to complete it if not half as good. In addition, peer grading or computer-grading which eliminate the role of teachers is seen as destroying the academic business model. Having personal experience MOOC as a participant, Kauza (2014) is not totally against MOOC but truly felt the lack of interaction with the instructor made the learning experience less fulfilling and of the opinion that MOOCs are more suitable for curious learners or certifications renewal rather as a credit-bearing higher education. The same opinion is shared by Syapin (2014) who also was not satisfied with her MOOC experience but agreed that it would help in education cost. Knox et. al. (2014) called for the engagement of participant's feedback in MOOC to explore and refine its design. With the emergence of less than a decade, MOOCs are still not fully mature and for sure will subject to refinement and correction. At the moment, MOOCs have not stopped but could be changing direction and morphing into next big thing (Krause, 2014). This study on MOOC conversational bot also plays the role of transforming MOOC into a more interactive platform.

## 2.3 Artificial Intelligence

The term "Artificial Intelligence" (AI) was first introduced at a workshop held at Dartmouth College in 1956 which saw the necessary to separate it from other fields of study (Russell and Norvig, 2010). AI is a branch of computer science which attempts to produce intelligent machine that could exhibit the behavior similar to human intelligence. In AI essence, it is an attempt to simulate the intelligence of human mind. According to Russell and Norvig (2010), AI system can be divided into four categories namely, thinking humanly, thinking rationally, acting humanly, and acting rationally as shown in Figure 2.1.

| **Thinking Humanly** | **Thinking Rationally** |
|---|---|
| "The exciting new effort to make computers think . . . *machines with minds,* in the full and literal sense." (Haugeland, 1985)<br><br>"[The automation of] activities that we associate with human thinking, activities such as decision-making, problem solving, learning . . ." (Bellman, 1978) | "The study of mental faculties through the use of computational models." (Charniak and McDermott, 1985)<br><br>"The study of the computations that make it possible to perceive, reason, and act." (Winston, 1992) |
| **Acting Humanly** | **Acting Rationally** |
| "The art of creating machines that perform functions that require intelligence when performed by people." (Kurzweil, 1990)<br><br>"The study of how to make computers do things at which, at the moment, people are better." (Rich and Knight, 1991) | "Computational Intelligence is the study of the design of intelligent agents." (Poole *et al.,* 1998)<br><br>"AI . . .is concerned with intelligent behaviour in artifacts." (Nilsson, 1998) |

Figure 2.1 Artificial Intelligence Categories (Russell and Norvig, 2010)

Today AI is used in many areas. Some of the examples are natural language understanding, speech recognition, planning and scheduling, vision, robotic, game playing, and expert system. There are many achievements or successful applications in AI history. For examples, computer program Deep Blue defeated human champion in chess game (Goodman and Keene, 1997), Asimo intelligent walking robot is able to interact with human using gestures and voice (Sakagam et. al., 2002), National Aeronautics and Space Administration (NASA)'s Mars exploration rovers, Spirit and Opportunity has autonomous navigation abilities (Mak, et. al., 2005), Google search by voice (Schalkwyk et. al., 2010), and IBM Watson defeated two human champions in America live television quiz show,

*Jeopardy!* (Tesauro et. al., 2012). The listings of the significant milestones in AI are summarized in Figure 2.2 chronologically.

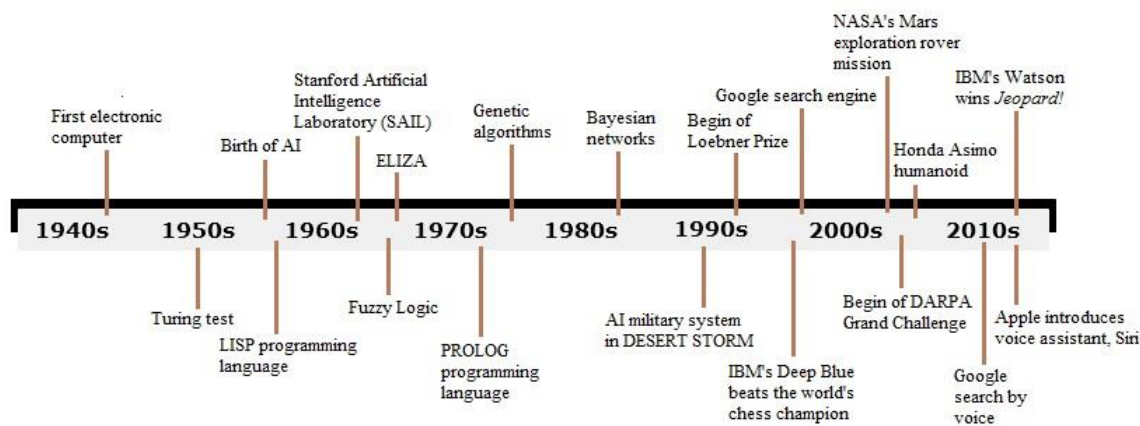

Figure 2.2 Artificial Intelligence Milestones

In practice, the progress of AI research was made possible with the creation of computer software and hardware. The software side supplied operating systems, programming languages, tools for writing modern programs, and databases. The advances of each generation of computer hardware have increased the computer's performance and capacity, resulted a more powerful but cheaper machine. These favorable developments have also made AI application become common and affordable to incorporate in consumer's products such as face recognition software in camera, floor-washing and vacuum-cleaning robots, self-adjusting washing machine for better clothes washing, self-adjusting air-conditioning system according to the ambience, and many more.

## 2.4    Turing Test, Loebner Prize and Chatterbox Challenge

Alan Turing was a highly influential pioneer in computer science as well as a brilliant mathematician. In 1950, through his famous article, "Computing Machinery and Intelligence", he proposed a behavioral intelligence test which he called it the "Imitation Game". His challenge on 'Can machines think?' in the paper brought the attention on the view of machine intelligence. The test which is more well-known as the Turing test (TT) now, suggested that a computer is considered "intelligent" if a human interrogator who posed some written questions is not able distinguish whether the written responses come from a human or a computer as illustrated in Figure 2.3.

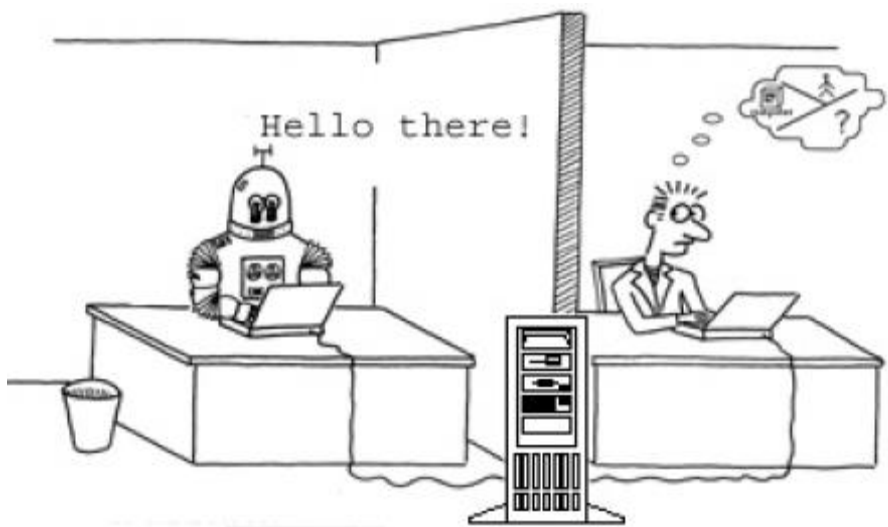

Figure 2.3 Turing Test (Goh, 2008)

Turing (1950) believed that in about fifty years from that time (which supposed to be year 2000), computers would have achieved the storage capacity of about $10^9$ which would see the test achieves the result of: "an average interrogator will not have more than 70 percent chance of making the right identification after five minutes of questioning". However, this notion proved to be wrong because as of today, the computer's storage have reached 1GB which he required and even exceeded that, but so far no intelligent computer program is able to achieve that level. This challenge set by Turing appeared harder than he expected.

TT brought lasting impact and great contribution in AI but it was not without objections and arguments which saw various literatures debated about it for more than half a century. Gunderson (1971), Searle (1980), Block (1981), French (1990), and Shieber (2006) have all questioned the sufficiency of TT as the criterion of intelligence which only relies on behavioral properties exhibited by a subject. Hayes and Ford (1995) argued that TT was not a sensible goal and considered it harmful to AI because AI researchers adopted it seriously as a long-range goal and motivation in the field which they think limiting AI's sights. However, there were also researchers that defended TT such as Moor (1976) and Leiber (1995) who argued that the critics had misunderstood or misinterpreted the real value of TT.

The Loebner Prize is the longest running annual contest based on TT but with some modification from the original TT (Shieber, 1994). Started since 1991, it was established by Dr. Hugh Loebner with the reward of $100,000 and a gold medal for the first program which able to respond indistinguishably from a human while each year the most human-like computer will win an annual cash prize and a bronze medal (Loebner, n.d.). Shieber (1994) has criticized that the competition has no clear purpose or any useful outcome but Loebner (1994) responded by defending the need of the competition as a tool to measure the state of the art in AI. Saygin et. al. (2003) agreed that the award in Loebner Prize could influence research in AI by making people interested in creating TT-passing programs. As of to date no one has come close to winning the gold medal which is passing the unrestricted TT.

The Chatterbox Challenge (CBC) is another annual TT-like contest which started in 2001. CBC has different category of awards such as 'best personality', 'conversational ability', 'most knowledgeable', 'best overall', etc. Unlike Loebner Prize which has specific venue for the competition, CBC is conducted over the Internet. This has raised the question on whether human could be intervening behind the system (Vallverdú et. al., 2010; Shah and Warwick, 2010). Due to this, CBC is seen carries less weight in conversational bot benchmark. Nevertheless, CBC is still a platform to test the state-of-the-art of conversational bots and comparison of technology.

## 2.5    Conversational Bots

One of the earliest and well-known works in conversational bot was the program called ELIZA. It was developed by Weizenbaum (1966) to play the role of a psychotherapist to a patient. This research was considered a breakthrough in AI at that time because it demonstrated a form of communication between man and computer in natural language. Although ELIZA was far from passing the real TT but being able to convince the interrogator to continue the conversation with the program led people to believe that it had demonstrated the responses in a Turing-like test. Figure 2.4 shows a sample dialogue of ELIZA from Weizenbaum (1966).

```
Patient     : Men are all alike.
ELIZA       : IN WHAT WAY
Patient     : They're always bugging us about something or other.
ELIZA       : CAN YOU THINK OF A SPECIFIC EXAMPLE
Patient     : Well, my boyfriend made me come here.
ELIZA       : YOUR BOYFRIEND MADE YOU COME HERE
Patient     : He says I'm depressed much of the time.
ELIZA       : I AM SORRY TO HEAR YOU ARE DEPRESSED
Patient     : It's true. I am unhappy.
ELIZA       : DO YOU THINK COMING HERE WILL HELP YOU NOT TO BE UNHAPPY
Patient     : I need some help, that much seems certain.
ELIZA       : WHAT WOULD IT MEAN TO YOU IF YOU GOT SOME HELP
Patient     : Perhaps I could learn to get along with my mother.
ELIZA       : TELL ME MORE ABOUT YOUR FAMILY
Patient     : My mother takes care of me.
ELIZA       : WHO ELSE IN YOUR FAMILY TAKES CARE OF YOU
Patient     : My father.
```

Figure 2.4 Sample Dialogue between ELIZA and Patient

As ELIZA was simulating a Rogerian psychotherapist, it could use the trick of rephrasing the patient's statements as questions and the patient will likely to open up his/her personal problems which will make the conversation continues. Weizenbaum was surprised that his program could also deceive normal people into believing it (Saygin et. al., 2003). There were suggestions went as far to replace psychotherapists with such kind of programs so that the therapy become cheap and could be done on a larger scale (Colby et. al, 1966).

ELIZA gave the impression of being more intelligent than it actually was. It has neither the understanding of English nor having the mind of a psychiatrist. It was implemented using pattern recognition according to decomposition rules and keywords substitutions according to reassemble rules where both were pre-defined scripts. If a keyword or phrase could not be identified, context-free remark such as "*please go on*" or "*does that have anything to do with*" the previous topic will be used. Figure 2.5 shows the decomposition rules and reassemble rules used in ELIZA.

```
keyword: your
 decomposition rule: * your *
  reassemble rule: Why are you concerned about
        my (2) ?
  reassemble rule: What about your own (2) ?
  reassemble rule: Really, my (2) ?
 decomposition rule: …
    reassemble rule: …

. . .
```

Example responses:

```
Patient    : What is your name?
Eliza      : What about your own name?
Patient    : Only your name, please!
Eliza      : Really, my name, please?
Patient    : Just tell me your name!
Eliza      : Why are you concerned about my name?
```

Figure 2.5 Decomposition Rules and Reassemble Rules with the '*' Represent Any Words

or Phrases

Inspired by ELIZA, Kenneth Mark Colby who was a psychiatrist wrote his own version of program to simulate the mind of a paranoid patient in the 70's. His model called PARRY had personality and emotional state such as 'fear', 'anger' and 'mistrust' which would determine its responses to the interviewer. This was done by having two separate modules, the parsing module and the interpretation-action module. The first module is used to interpret the meaning of the input in natural language and it is passed to the second module which keeps the current state of mind to decide the linguistic actions to respond (Colby, 1975).

PARRY was tested in Turing-like test to determine whether experienced psychiatrists could distinguish between the program and a real human patient (Colby, 1981). The overall test results showed that psychiatric judges were unable to make the relevant distinctions between PARRY and real paranoid patients. This was due to the fact that paranoid people have irrational and obsessive behavior, therefore, inappropriate responses seemed reasonable. It may be simpler to simulate paranoid behavior but there is still clever logic that made PARRY simulates effectively. Figure 2.6 shows the conversation between PARRY and a human psychiatrist where it answered well and attempted to introduce its own topic.

```
Interviewer:    Good afternoon. My name is Doctor Jones.
PARRY:          Pleased to meet you.
Interviewer:    How are you?
PARRY:          All right I guess.
Interviewer:    Could you tell me your name?
PARRY:          Pat Smith.
Interviewer:    How old are you?
PARRY:          28.
Interviewer:    I thought you might have some problems.
PARRY:          I've been upset lately.
Interviewer:    By what?
PARRY:          People get on my nerves sometimes.
Interviewer:    Tell me more.
PARRY:          Do you know anything about bookies?
Interviewer:    No. What about bookies?
PARRY:          A bookie didn't pay me once.
Interviewer:    When did this happen?
PARRY:          A couple of months ago.
```

Figure 2.6 Conversations between PARRY and Human (Colby, 1981)

Interestingly, ELIZA and PARRY were put on conversation for a demonstration in 1973. Figure 2.7 shows part of their conversation. The beginning of their conversation was reasonable but as the conversation went on, both were trying to repeat what the other said. This would not be difficult for someone in TT to decide whether it is a program or a human.

```
ELIZA: Good Evening. Tell me your problems. Please terminate input with a
period or a question mark.
PARRY: People get on my nerves sometimes.
ELIZA: I am not sure I understand you fully.
PARRY: You should pay more attention.
ELIZA: Suppose you should pay more attention.
PARRY: You're entitled to your own opinion.
ELIZA: What makes you think I am entitled to my own opinion?
PARRY: It bothers me just to be around people in general.
ELIZA: Please go on.
[Conversation continues...]
PARRY: I don't understand your motives.
ELIZA: Why don't you understand my motives?
PARRY: Shouldn't I be?
ELIZA: Why do you say that just now?
PARRY: That's just the way it is.
ELIZA: Suppose that's just the way it were not.
PARRY: So?
ELIZA: I am not sure I understand you fully.
PARRY: You harp on the same questions too much.
ELIZA: What does that similarity suggest to you?
[Conversation continues...]
```

Figure 2.7 Conversations between PARRY and ELIZA (Cerf, 1973)

In 1990 Dr. Hugh Loebner founded the Loebner Prize with the aims of promoting AI research and formalizing the Turing test. The competition offers the Grand Prize of $100,000 and a gold medal for the first program which able to respond indistinguishably from a human, while each year the most human-like computer will win an annual cash prize and a bronze medal. Since its inception, the competition has attracted the development of various conversational bots to win the prize each year. The winners of the Loebner Prize from 1991 to present are listed in Table 2.1. Among these winners, the notable ones are those that won more than once such as Albert One, ALICE, and Jabberwacky.

Table 2.1 The Winners of Loebner Prize from 1991 to 2014 (Loebner, n.d)

| Year | Author | Conversation Bot | Year | Author | Conversation Bot |
|------|--------|------------------|------|--------|------------------|
| 1991 | Joseph Weintraub | PC Therapist | 2003 | Jorgen Pirner | Jabberwock |
| 1992 | Joseph Weintraub | PC Professor | 2004 | Richard Wallace | ALICE |
| 1993 | Joseph Weintraub | PC Politician | 2005 | Rollo Carpenter | George (Jabberwacky) |
| 1994 | Thomas Whalen | TIPS | 2006 | Rollo Carpenter | Joan (Jabberwacky) |
| 1995 | Joseph Weintraub | PC Therapist | 2007 | Robert Medeksza | Ultra Hal |
| 1996 | Jason Hutchens | Hex | 2008 | Fred Roberts | Elbot |
| 1997 | David Levy | Converse | 2009 | David Levy | Do-Much-More |
| 1998 | Robby Garner | Albert One | 2010 | Bruce Wilcox | Suzette |
| 1999 | Robby Garner | Albert One | 2011 | Bruce Wilcox | Rosette |
| 2000 | Richard Wallace | ALICE | 2012 | Mohan Embar | Chip Vivant |
| 2001 | Richard Wallace | ALICE | 2013 | Steve Worswick | Mitsuku |
| 2002 | Kevin Copple | Ella | 2014 | Bruce Wilcox | Rose |

ALICE (Artificial Linguistic Internet Computer Entity) is a conversational bot developed by Dr. Richard Wallace and won the Loebner Prize in 2000, 2001, and 2004. It was built with Artificial Intelligence Markup Language (AIML) which is a of Extensible Mark-up Language (XML)-based language. AIML was created by Dr. Richard Wallace and the Alicebot open source community between 1995 and 2000. ALICE uses simple stimulus-response technique to return matching response to the user, similar to those used by ELIZA but it has more categories of knowledge compare to ELIZA. The stimulus-response structure of ALICE resembles Case-Based Reasoning (CBR) paradigm (Wallace, 2009). Figure 2.8 shows part of the conversation between the judge and ALICE during the Loebner Prize competition.

| Alice: | Hello, my name is Alice and I am the human. |
| Judge 2: | and what do you do in life as we know it? |
| Alice: | I like to meet new people online. |
| Judge 2: | that sounds a bit scary |
| Alice: | It sounds like it to me too. |
| Judge 2: | so you're either brave or foolhardy |
| Alice: | Interesting deduction. Do you mind if I tell other people you say I am either brave or foolhardy. |
| Judge 2: | sure, it will make me sound Aristotilian, which I don't mind at all |
| Alice: | That is only what you think. |
| Judge 2: | no, I'm saying aristotle thought it first, actually |
| Alice: | Are you a man or a woman? |
| Judge 2: | can you guess? |
| Alice: | Guess what? |
| Judge 2: | guess whether I'm a man or woman? |
| Alice: | I can't guess. Tell me. |
| Judge 2: | man |
| *[Conversation continues...]* | |

Figure 2.8 Conversations between ALICE and Judge During the Loebner Prize

Competition (Loebner, 2004)

ALICE can be added with new responses using AIML content. This AIML content can be created by the botmaster with anticipatory approach where he/she will try to guess all or most of the likely ways clients might ask the same question, or express the same statement. Another approach of creating AIML content would be based on a backward-looking log file analysis. In this method, the botmaster will check on the logged conversations and look for "incorrect replies" in the dialogue. This will be followed by rewriting new categories for those queries. The backward looking approach is in accordance with Zipf's Law (Zipf, 1949) where a query uttered by a person will most likely be uttered by others again later. This will ensure that the most commonly uttered sentences are taken care of (Wallace, 2009).

The set of AIML scripts for ALICE has been made available for free download at Google Code. It is known as the Annotated ALICE AIML files (AAA) comprising the brain of the award winning conversation bot. The AAA files which mostly compatible with all AIML 1.0.1 enable botmasters to easily clone ALICE's brain and thus reduce the huge efforts to create and edit the AIML contents (Wallace, 2011). AIML technology has spread widely due to the availability of AAA files and AIML under General Public License (GPL). Beside this, many AIML interpreters has been made available in different programming language such as Lisp, Java, PHP, Ruby, C++ and many more. Each interpreter has its own name, for example, Program D is the AIML interpreter for Java and Program O is the interpreter for PHP. Table 2.2 lists all the name of the interpreters and their programming language.

Table 2.2 AIML Interpreter (ALICE AI Foundation, n.d.)

| Interpreter | Language |
|---|---|
| Program Z | Lisp |
| Chatterbean | Java |
| Program D | Java |
| Program O | PHP |
| Program E | PHP |
| Program R | Ruby |
| Program V | Perl |
| Program P | Pascal |
| Program Y | Python |
| Program # | .NET |
| Program Q | C++ |
| libaiml | C++ |
| J-Alice | C++ |
| RebeccaAIML | Object Oriented C++ |
| Program M | SETL |

ALICE AI Foundation (n.d.) claimed that AIML has captured more than 80% of the world market for conversational bot technology. Indeed, due to the availability of AIML under GPL, there are many conversational bot implemented based on AIML technology whether in education, government, medical, entertainment or commercial. Mahapatra et. al. (2012) created AIML-based chatter bot as interface tasks in E-Government interactive support systems with the aim to address audience who are computer illiterate. De Gasperis (2010) used AIML to build the knowledge of an Italian chatter bot. Persona-AIML (Galvao et. al., 2004) is the architecture for developing chatter bots with personality which favors the reuse of AIML bases. Humorist Bot (Augello et. al., 2008) is an AIML conversational agent with humoristic features which was embedded as a Yahoo instant messaging contact. There are also researches which integrate AIML as part of their architecture or model. Among these works are multimodal human-robot interaction with AIML chatter bot system (Tan et. al., 2012), and Artificial Intelligent Neural-network Identity (AINI) architecture (Goh, et. al., 2007) in human-computer communication system.

In education field, CHAtteR Learning Interface Entity (CHARLIE) is the AIML chatter bot responsible for communication between the INtelligent Educational System (INES) and the students in natural language (Mikic et. al, 2009). T-Bot is the tutor bot which able to provide domain specific answers in natural language according to course contents when requested by students while Q-Bot is an evaluation bot which keep track and supervises the student progress through personalized questionnaires. Both are implemented with AIML in e-learning platform (Mikic et. al., 2008). University Faqbot (UFAQbot) is another example of AIML chat bot for education purpose. UFAQbot served as undergraduate advisor in student information desk to help students with admission and course information (Ghose and Barua, 2013).

Despite the popularity of AIML, it is not without its limitation. For example, the use of depth-first search algorithm in AIML to retrieve the responses does not optimize the results because it only finds the first available solution, and there is no spell-check function in AIML (Lu, et. al., 2006). This has prompted the researches to enhance AIML-based conversational bot. Among them is AZ-ALICE which built on ALICE Program D as the chat engine but customized with perl script that allows users to evaluate and suggest new patterns (Schumaker et. al., 2006). Intentional AIML (iAIML) is the mechanism to treat intention information in AIML chatter bots by incorporating intentions and rules used in sentence interpretation and generation (Neves et. al., 2006). Torrey et. al. (2006) improved their robot dialogue interaction by replacing AIML's depth-first search with an A* search, and added priorities and a system to find the best match to a question. Weerawarna et. al. (2011) overcome weaknesses in existing AIML chatter bots by introducing new scripting language called CyberMate Scripting Language (CSL) which comes with CSL integrated development environment (IDE) to develop the scripts and CSL parser to compile CSL scripts to AIML.

Another Loebner Prize winner, Jabberwacky was created by British programmer Rollo Carpenter. Jabberwacky was also using contextual pattern matching techniques but what makes it unique was that it learnt from conversation which models the way humans learn language, facts, context and rules. This makes it able to learn foreign language, slang English, word games, and jokes through conversation. Without hard-coded rules and long term memory, Jabberwacky relies entirely on the feedback from the conversation logs which it recorded for the responses. Jabberwacky's incorporation of AI purely from conversation only suitable for general intelligence and character but not meant for business or commercial purpose. Figure 2.9 shows a snippet of conversation between Jabberwacky and a human user.

| User: | Do you know who Alan Turing is? |
|---|---|
| **Jabberwacky**: | Isn't he the inventor of the Turing Test? |
| User: | Correct. Did you pass the test? |
| **Jabberwacky**: | Yes. I correctly identified you as a bot. |
| User: | Right. Do you gain intelligance by speaking to humans? |
| **Jabberwacky**: | Yes. |
| User: | Are you human? |
| **Jabberwacky**: | Human? |
| User: | Human. |
| **Jabberwacky**: | And so am I. |
| User: | So we're both humans? |
| **Jabberwacky**: | I am human, you are computer who talks to the website which is maintained by humans. |
| User: | Actually, I am human. You are a computer who talks to intelligent people like myself. |
| **Jabberwacky**: | I'm not a computer but I like men. |
| User: | So you are female? |
| **Jabberwacky**: | Yeah. I'm a female. |
| User: | Ah. So am I. |
| *[Conversation continues....]* | |

Figure 2.9 Conversations between Jabberwacky and a Human User (Carpenter, 2009)

In commercial environment, conversational bots usually act as the virtual customer service agent which able to provide 24/7 services. The benefits include offering friendly, tireless services, and the ability to answer the most common questions. Anna is an example of commercial conversational bot. She is the female virtual assistant for IKEA, the Swedish furniture retailers. Anna can be seen as a 2D animated image wearing official company's theme of yellow shirt with blue collar at certain countries IKEA's website. She accepts typed input and reply in text in many languages. The role of Anna is mainly to guide users in IKEA-related enquiries such as products, prices, sizes, delivery, etc. Anna will also open related pages in a browser window to facilitate user who are looking for information. There is not much information published regarding the technique used behind Anna's creation because it is developed by a commercial company called Artificial Solutions. However, Shah and Pavlika (2005) mentioned that Anna was based on technology of Elbot which is the winner of 2008 Loebner Prize and runner up in year 2003.

## 2.6 Artificial Intelligence Mark-up Language (AIML)

AIML was designed by Dr. Richard Wallace and Alicebot free software community to create stimulus-response chat bots. It is a derivative of XML with a specific structure. As an XML-compliant language, AIML also respects the grammar specification in XML such as the syntax of comments, white space handling, end-of-line handling, the need of closing tag, and so on. An AIML file should be saved with ".aiml" extension, for example: "bot.aiml" and it must enclose "<aiml>" tag in file. The "<aiml>" tag may contain the AIML version as shown in line 1 of Figure 2.10 to avoid confusion because there is more than one version such as 1.0, 1.0.1 and the latest is 2.0 which its specification is still under draft (Wallace, 2014).

The basic unit of knowledge in AIML is called "category" and it is written as "<category>" in AIML file. Usually the more categories an AIML file has, the more knowledge it contains. Each "category" is followed with input or question known as "pattern", written as "<pattern>" and the answer is matched with one or more output called "template" which is written as "<template>". The AIML pattern is consisting of only of words, spaces, and the wildcard symbols '_' and '*'. Beside this, there are two types of optional context called "that" (<that> tag) and "topic" (<topic> tag) which are part of the AIML main elements but not compulsory to be present. The "<that>" tag tells the system to refer to the last sentence presented by the chatterbot while the "<topic>" tag is used to organize subjects or topics that only applied to an active topic variable. The basic units of AIML which are mandatory in AIML file are shown in Figure 2.10.

```
1  <aiml version="1.0.1">
2  <category>
3  <pattern> HELLO</pattern>
4  <template>
5  Hello there! How are you?
6  </template>
7  </category>
8  </aiml>
```

Figure 2.10 Basic Units in AIML

In the templates, there are additional tags that can be used according to the needs of the chat bot. Table 2.3 summarizes some of the commonly-used tags and their functions. In addition, AIML also supports the interface of other languages and systems. The "<system>" tag executes any program accessible as an operating system shell command, and inserts the results in the reply. On the other hand, the "<javascript>" tag allows the execution of JavaScript code inside the templates.

Table 2.3 List of Some of the Tags Used in Template and Their Functions (ALICE AI Foundation, 2011)

| Tag | Function |
| --- | --- |
| <think> | The <think> tag is located in the template section and allows the chatterbot to process and compute information about the matched pattern without display to the user. This tag is used for data processing, conditional statements and tests that should not be visible to user. |
| <srai> | The <srai> tag can link multiple categories together. It can be used for recursion, giving one single template section for different patterns that have the same meaning (are synonyms or close), it can be used to map very similar specific patterns to a more generic one (when the bot needs to provide a generic answer), and it can be used as a run-time synonym replacer for words not matched in the invoked substitution file (which cannot be changed at run time). |
| <set> | The <set> tag is used to create variables. |
| <get> | The <get> tag returns the value stored by <set> tag. |
| <star index="n"/> | The <star index="n"/> tag captures a particular text fragment, contained in the user input sentence. The index n indicates the phrase component that will be mapped and captured. |
| <condition> | The <condition> tag is used when there is a list of possible answers to be presented to the user, and the choice of the most appropriate responses relies on the analysis of a particular variable that was updated during the conversation between the user and the chatterbot. |
| <random> | The <random> tag is used to respond to a user input in different ways. |
| <li> | General list item of what the chatterbot's response could be. It is used with <random> and <condition> tag. |
| <bot> | Allows developer to define some chatterbot's properties that can be seen by the user during the conversation. |

An AIML interpreter will store all the categories in a tree using Graphmaster object. The goal is to make pattern matching efficient by not searching in a brute force manner. The Graphmaster works by looking for the section that matches the first letter or word and skip to another section that contains the set of the next letter or word. The process continues until the word or phrase is found. This is done with the help of a set of nodes called Nodemappers. The Nodemappers map branches from each node where the branches are either single words or wildcards. The matching steps in Graphmaster begin with matching the "_" listings first followed by alphabetical listings next. If no matches are found, the patterns beginning with "*" will be searched as shown in Figure 2.11.

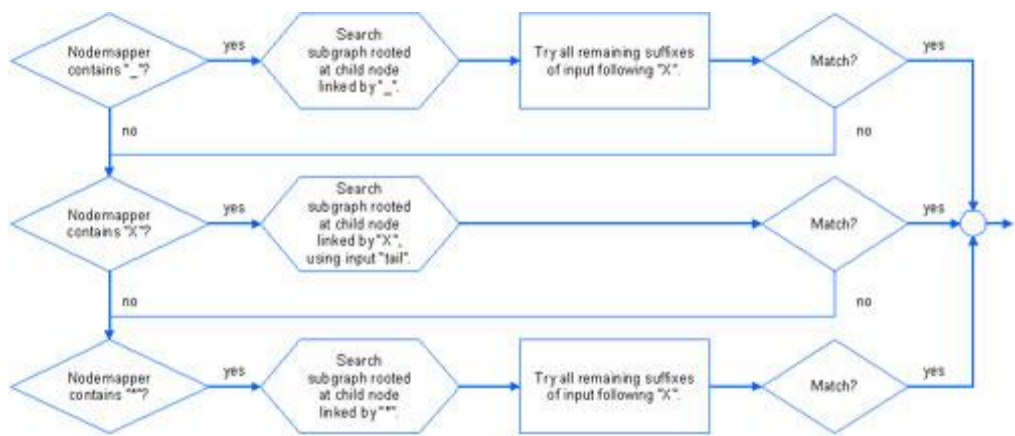

Figure 2.11 Matching Algorithm in Graphmaster (ALICE AI Foundation, 2011)

The Graphmaster is a metaphor of a file system hierarchy whether in Windows, Unix or Mac system. It is analogous to a file's path name such as "C:\My Documents\My Pictures\photo.jpg" that represents the file's location within the system. The AIML pattern of "I LIKE TO *" is equivalent to "g:\I\LIKE\TO\star" which start with the "g:\I\" folder followed by "g:\I\LIKE\" subfolder that contains a single file called "template.txt" which is the template of AIML (Wallace, 2007).

**2.7    Web Speech Application Programming Language (API)**

Web Speech API provides JavaScript interface for developer to integrate speech recognition and speech synthesis into web applications. Speech recognition is an alternative input method which works by accepting speech input and converts them to text. Speech input has several benefits, for example, it enhances accessibility for visually-impaired users and offers as a hands-free input method (Adorf, 2013). On the other hand, speech synthesis is the act of transforming text to speech. Speech synthesis is incorporated in Text-to-Speech (TTS) applications to provide better user experience.

Even though Web Speech API specification is included in HyperText Markup Language 5 (HTML5), so far only Google Chrome and Apple Safari browsers supported them (Walther, 2015). One of the drawbacks of Web Speech API is user have to grant the application permission to use the microphone for speech recognition purpose. If the page is not hosted on Hypertext Transfer Protocol Secure (HTTPS), the permission needs to be granted for each utterance. This can be overcome by applying Secure Sockets Layer (SSL) certificate to the website in order to establish an encrypted link between the website and client and this will be reflected with the HTTPS uniform resource identifier (URL). By using HTTPS hosting, the user only needs to grant the permission once even if he/she returns to the website multiple times.

**2.8    Summary**

This chapter gives an overview on MOOCs from a few perspectives such as types of MOOCs, advantages, and challenges. One of the issues in MOOCs which is the lack of interactivity with the instructor has brought conversational bot into the picture to fill in this gap. Conversational bot which is considered part of AI's application is always being measured in Turing test envisioned by Alan Turing. Although some may have objected TT as the focus in AI research, the Loebner Prize and Chatterbox Challenge competition has kept it alive in the past 20 over years with annual prize to attract participants. As of to date, the grand prize of Loebner Prize remains unclaimed but the techniques in conversational bot has evolved with AIML from ALICE as one of the impactful outcomes. AIML may not be perfectly intelligent technology but its capability to deliver appropriate responses in domain-restricted areas (Schumaker et. al, 2006; Mikic et. al., 2008; Tan et. al, 2012; Ghose and Barua, 2013) is convincing enough to be the design behind MOOC-bot.

# CHAPTER 3

# RESEARCH METHODOLOGY

## 3.1 Introduction

Research methodology is important to solve the research problem systematically. This chapter discusses the methodology and design used in this research with the aim of building a MOOC-bot prototype. Having reviewed the AIML technology in the previous chapter, it will now be applied in MOOC-bot to play the role of conversational bot that can respond to course-related enquiries. The prototype of MOOC-bot will have the domain knowledge of "MITI 5113 Artificial Intelligence" course taught in UTeM.

## 3.2 Type of Research Method

This study is based on qualitative research because it cannot be easily quantified in numerical data as needed in quantitative research. Initially largely used in psychology and social sciences, the use of qualitative research has been growing in the past decades due to the tendencies for empirical research in tandem with modernism (Flick, 2009). Qualitative research is an approach which involves inductive data analysis and data interpretation from the researcher (Creswell, 2013). For conversational bot's research, there are many of them conducted in qualitative method (Benotti et. al., 2014; Ghose and Barua, 2013; Dybala et.

al., 2009) because Turing test which is popular among researchers in this field is also based on qualitative assessment (Browne, 1991).

## 3.3    Research Design

In order to build MOOC-bot, the software development model of prototype will be used to ensure a rapid development in a less expensive way. Prototype is a direct method to provide the picture of the proposed system to the users. According to Pressman and Maxim (2014), prototype that has been evaluated by user can be used to refine the requirements of the software.  Prototype can be adapted to fulfill user's requirements and at the same time the developer can have better understanding of the requirements based on the feedback. The changes and suggestions done in prototype stage can avoid sudden change after the completion of the system.

Prototype methodology has two strategies: requirement prototype which is also known as throwaway prototype, and evolutionary prototype. The development of MOOC-bot would be using evolutionary prototype because instead of discarding the prototype, it will be refine to become a real system. The prototype which has been verified will be used as part of the final system. Figure 3.1 shows the activities involved in evolutionary prototype.

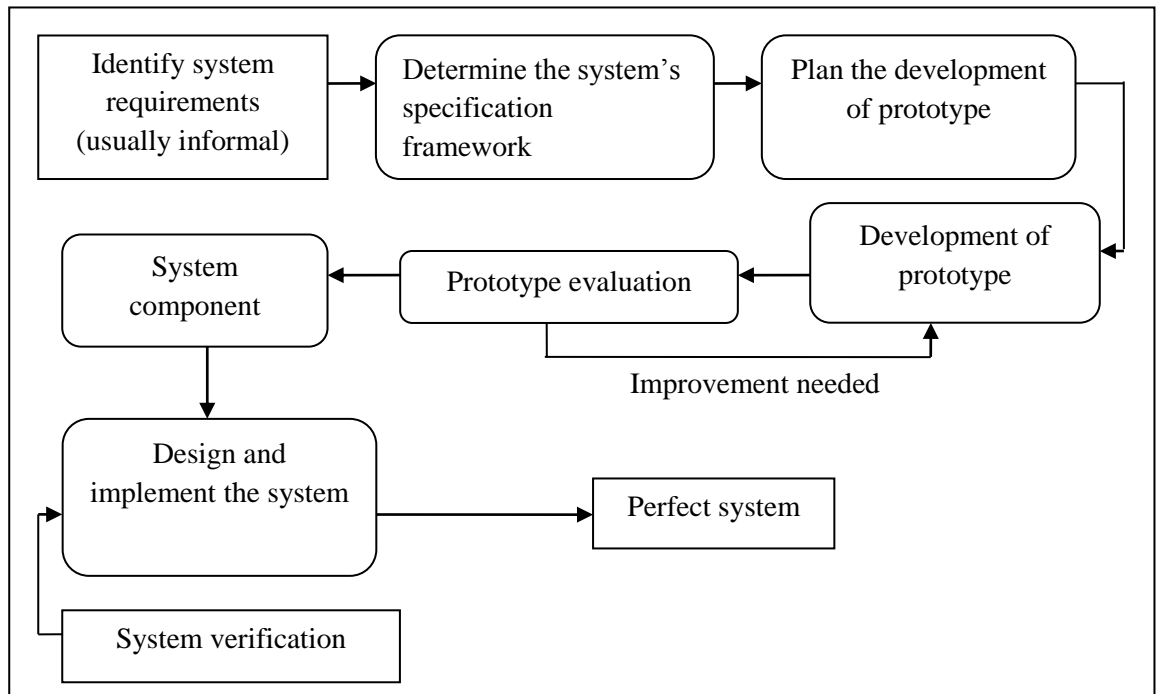

Figure 3.1 Evolutionary Prototype Model

Based on Figure 3.1, the first activity in prototype system development is to identify the system requirements. This activity in MOOC-bot would be identifying the syllabus and lectures given in "Artificial Intelligence". After that, the specification framework will be determined to produce MOOC-bot prototype. This prototype will be evaluated by lecturer and users repetitively until it is able to fulfill the questions of majority users. The next activity is to create the design and implement the system. After the system is completed, it will be verified to ensure that the system is perfect.

Generally, evolutionary prototype strategy consists of five phases as described below. All these five phases will be used in MOOC-bot development:

i)    Planning Phase

This phase will determine what type of methodology to be used and identify what are the current problems.

ii)   Analysis Phase

This phase will identify and produce the end-user specification.

iii)     Design Phase

This phase will produce the prototype design based on the end user specification.

iv)     Prototype Development Phase

This phase will involve the development of the prototype.

v)      Prototype Testing Phase

This phase will test the developed prototype and show to the users.

### 3.3.1   Planning Phase

In this phase, the selection of which methodology to be used will be done. Evolutionary prototype methodology has been chosen because of its advantage to provide a working model to the end users to better understand how MOOC-bot would work before the complete system. The existing problems of MOOC are identified and described in Chapter 1.

### 3.3.2   Analysis Phase

This phase will study the specification of the system to be developed based on the problems that have been identified in planning phase such as the need to serve huge number of learners and to provide 24-hour service for the convenience of students from different time zones. The knowledge needed for this prototype such as course contents of "Artificial Intelligence" and basic general knowledge can be obtained from:

i)      Lecture Note

The lecture notes are provided by the instructor of this course.

ii)    Book

The main reference book of the course is "Artificial Intelligence: A Modern Approach" (Russell and Novig, 2010).

iii)   Annotated ALICE AIML (AAA) Files

The AAA files which provided by ALICE AI Foundation contain a lot general knowledge which should be sufficient for MOOC-bot to handle general conversation.

### 3.3.3   Design Phase

After identifying the user specifications in analysis phase, the design of the prototype can be produced. The design is MOOC-bot's system architecture which consists of chat interface, knowledge base, AIML interpreter and MOOCs website. The detailed description of the system architecture is explained in 3.4.

### 3.3.4   Prototype Development Phase

After completed the prototype design, the actual prototype will be developed. The first prototype that fulfills the basic requirements will be developed and shown to the course instructor. MOOC-bot prototype will be integrated with the MOOC website. This prototype will be able to gather results from the usage of users at the website by referring to the conversation logs saved in the database. Subsequently, the second prototype will be done by improving the first prototype knowledge base by analyzing the conversation logs. The second prototype will repeat the same process as the first prototype. The last prototype which is able to answer most of the questions rationally will be used as the real system.

The development is done in Windows system and can be tested using local web server-client environment. This can be done by installing *Apache* web server in Windows. *WampServer* package which can be downloaded from the Internet can automatically create *Apache*, *Hypertext Preprocessor* (PHP) and *MySQL* database in Windows when installed. It also comes with *PhpMyAdmin* to manage the databases conveniently. Once *WampServer* is installed, the root directory "*www*" will be automatically created at "*c:\wamp\www*". The MOOC-bot application will be placed in the root directory and can be accessed using "*localhost*". The software and hardware needed for MOOC-bot development are summarized in Table 3.1 and Table 3.2 respectively.

Table 3.1 Software Involved in MOOC-bot Development

| No | Software | Function |
|----|----------|----------|
| 1 | Microsoft Windows 7 | As server. |
| 2 | Apache Web Server | As web server. |
| 3 | Hypertext Preprocessor (PHP) | Programming language of the system. |
| 4 | Google Chrome | As web client. |
| 5 | MySQL | Relational database management system (RDBMS). |
| 6 | Macromedia Dreamweaver 8 | For interface design and PHP programming. |
| 7 | Microsoft Word 2010 | For project documentation. |
| 8 | Notepad++ | To edit AIML file. |
| 9 | WampServer 2.5 | Installer for latest releases of Apache, MySQL and PHP. |

Table 3.2 Hardware used in MOOC-bot Development

| No | Hardware | Usage |
|----|----------|-------|
| 1 | Laptop | Work station to develop MOOC-bot. |
| 2 | Storage devices (External hard disk/Pendrive) | To store all the materials and reports as a backup for the project. |

### 3.3.5 Prototype Testing Phase

In prototype testing, the test is done to ensure the logic is correct and able to produce the expected result. In this case, MOOC-bot should be able to respond to enquiries related to the domain knowledge which is the course's contents and some of the general questions. This test can be done by using some of the past questions posed by judges in conversational bot's contest such as Loebner Prize and Chatterbox Challenge Competition.

### 3.4 System Architecture

MOOC-bot is integrated into MOOCs website as its chat interface. Just like any web application, it works using client-server concept. The MOOCs website and AIML interpreter are hosted at *Apache* web server while the database and knowledge base are using MySQL that runs at the server. The web server works by translating PHP codes into HyperText Markup Language (HTML) while the web client such as Internet Explorer, Google Chrome, and etc. will translate the client script such as JavaScript before display to the users. If the user is using Google Chrome browser and chooses to use speech recognition, the Web Speech API will convert speech to text before sending to AIML interpreter. The result will be displayed in text and speech. The general database is used to save MOOCs website related information such as user's profile, course's subscription and so on.

When a user poses questions to MOOC-bot's chat interface, the question will be processed by AIML interpreter by referring to the knowledge base which contains the course's knowledge, the course's frequent asked questions (FAQ) and the general knowledge that comes from AAA files. The botmaster can constantly maintain the knowledge base by adding the knowledge in AIML file format. Figure 3.2 illustrates the overview of the system architecture.

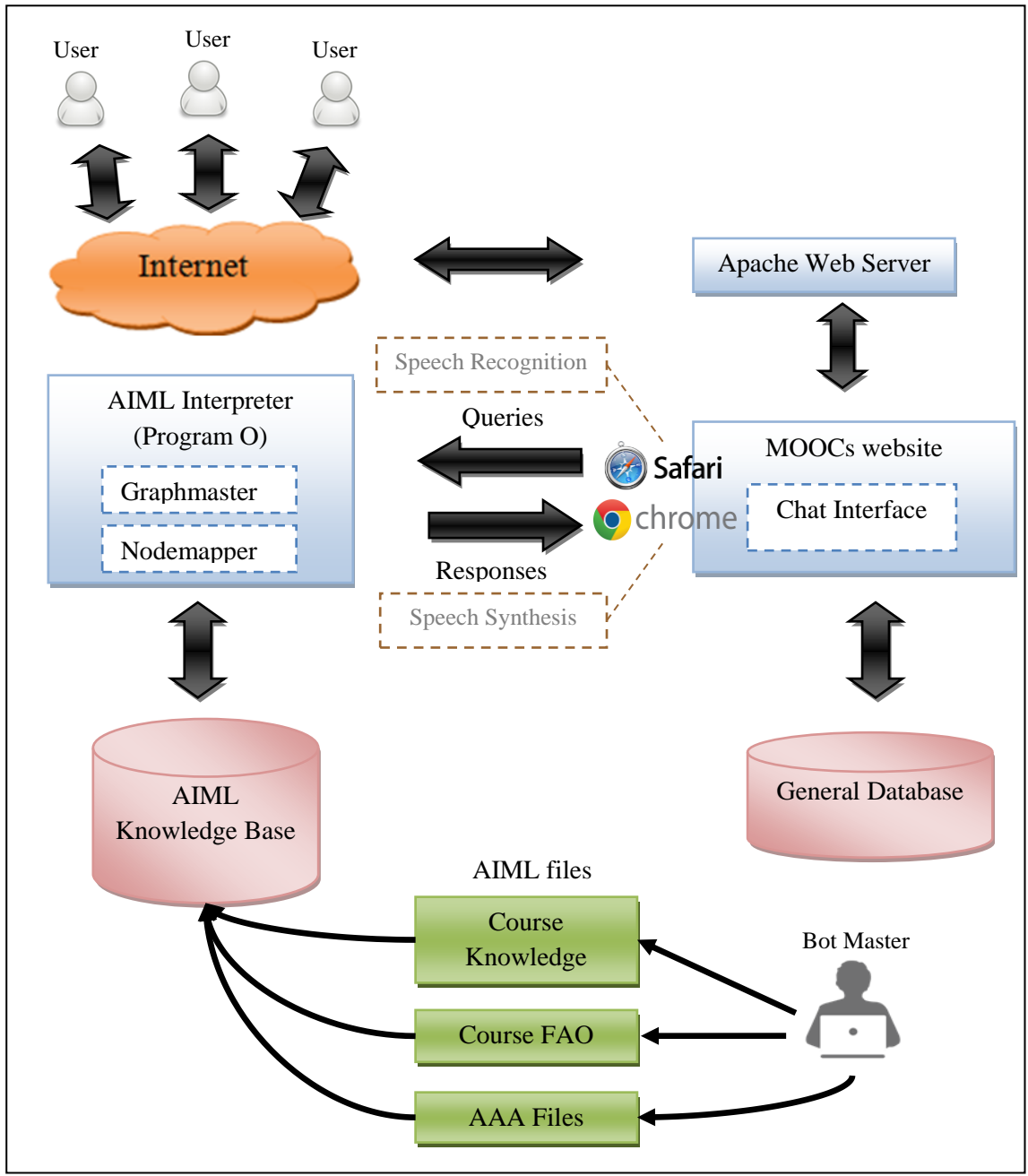

Figure 3.2 MOOC-bot's System Architecture

### 3.4.1   Chat Interface

MOOC-bot's chat interface is integrated with MOOCs website. The chat interface is located at the bottom of the website as shown in Figure 3.3. The function of chat interface is to receive text or speech input from the user and pass it to the AIML interpreter. For voice recognition, user needs to press the microphone button and grant permission for the application to receive voice input. The results from the interpreter will be displayed and read out to the users if the speaker is turned on. Figure 3.4 shows the chat interface by itself.

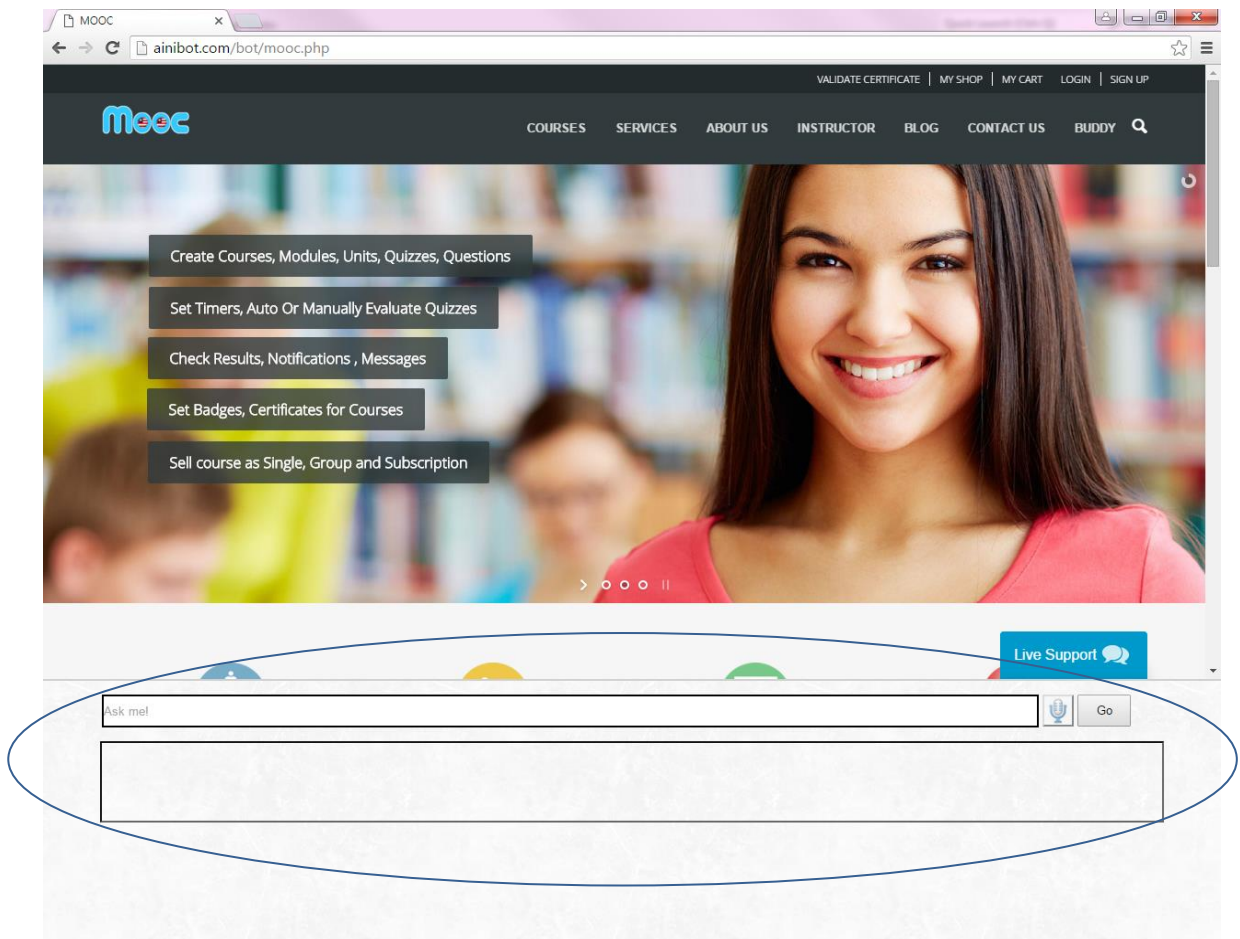

Figure 3.3 Chat Interface in MOOCs Website

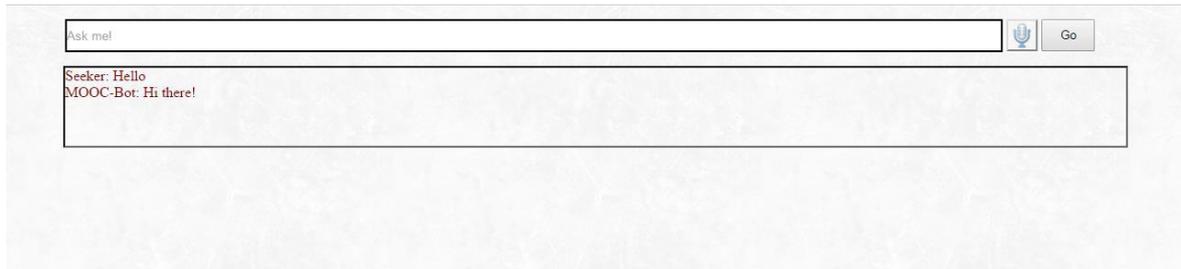

Figure 3.4 MOOC-bot Chat Interface

### 3.4.2 Knowledge Base

The knowledge of MOOC-bot consists of the course's knowledge, the course's FAQ and the general knowledge that comes from AAA files. The course's knowledge is "Artificial Intelligence" offered by UTeM and the syllabus is shown in Table 3.3. The course's FAQ can be added into the knowledge base using backward-looking log file analysis by the botmaster. This is done by reading the logged conversations and look for frequently asked questions in the dialogue and add the responses to the knowledge base using AIML file. This will ensure that the most common enquiries are taken care of. The AAA files comprise the brain of the award winning conversation bot ALICE. The availability of these files for free download at Google Code makes the job of cloning ALICE's brain easy.

Table 3.3 Syllabus of "Artificial Intelligence"

| Lecture | Lesson |
|---------|--------|
| Chapter 1 | Introduction to Artificial Intelligence |
| Chapter 2 | Intelligent Agents |
| Chapter 3 | Problem Solving by Searching |
| Chapter 4 | Adversarial Search |
| Chapter 5 | Logical Agents And Fuzzy Logic |
| Chapter 6 | Knowledge Representation |

An AIML file can be created using any text editor such as *Notepad++* and saved with ".aiml" extension. It is a derivation of an XML file with a specific structure. The course's knowledge and course's FAQ are written in AIML files and save into the knowledge base. An example of AIML file content for the course's knowledge is shown in Figure 3.5.

```
<aiml version="1.01">
<category>
<pattern>WHAT IS AI</pattern>
<template> AI is a branch of computer science which attempts to
produce intelligent machine that could exhibit the behavior
similar to human intelligence. In AI essence, it is an attempt
to simulate the intelligence of human mind.
 </template>
</category>

<category>
<pattern>WHAT IS ARTIFICIAL INTELLIGENCE</pattern>
<template><srai> WHAT IS AI </srai> </template>
</category>
```

Figure 3.5 Example of an AIML File for Course's Knowledge

### 3.4.3 AIML Interpreter

MOOC-bot is using *Program O* as its AIML interpreter. *Program O* is an AIML engine written in PHP and uses *MySQL* database to store the bot's information, including the AIML files used to generate the bot's responses. It is available for free download at https://github.com/Program-O/Program-O. *Program O* is suitable for MOOC-bot because it is implemented in web environment which able to be executed independently and thus made it possible to be integrated to the MOOCs website. The text or speech input from the chat interface will be processed by the interpreter by performing pattern matching with the AIML content in the knowledge base using depth-first search. The first matching response will be returned to the chat interface. The support of recursion in AIML makes it simple but yet powerful in pattern matching.

### 3.4.4   MOOCs Website

The MOOCs website is the initiative of UTeM to provide massive open online courses for students. For the prototype, it is hosted at http://ainibot.com/bot/mooc.php. A student can register for a course, download lecture notes, take quiz, submit assignments, and participate in a forum from here. The progress of a student can also be tracked and monitored by the instructor. The integration of MOOC-bot's chat interface in this website complements the role of the website.

### 3.5   Summary

This chapter gives the overview of the research method, research design, and system architecture. Qualitative research approach discussed in this study is in line with the research objectives which are qualitative in nature. The prototype of MOOC-bot is part of the research design to provide user experience in order to assess its performance in qualitative way. The system architecture presented in this chapter is essential to build the conversational bot for MOOCs. The chat interface, knowledge base, and AIML interpreter are the most important components in MOOC-bot's design.

# CHAPTER 4

# IMPLEMENTATION

## 4.1    Introduction

This chapter discusses the implementation of MOOC-bot prototype. Having reviewed the architecture in the previous chapter, MOOC-bot will be developed based on the design which includes chat interface, knowledge base, and AIML interpreter. In order to prepare for the development, the environment needs to be setup up locally and the relevant software needs to be installed. Subsequently, the MOOC-bot's knowledge can be built up using the interpreter's interface. With the appropriate environment for development, the prototype can be developed and used for performance assessment in the next chapter.

## 4.2    Software Installation

MOOC-bot is a web-based conversational bot using *Program O* interpreter. *Program O* is an AIML interpreter written in PHP and uses *MySQL* database to store the data. In order to create server-client environment in development's computer, *Apache* web server needs to be installed. *WampServer* package which is free for download at its website provides the convenience of automatically create *Apache*, PHP and *MySQL* database in Windows when installed. It also comes with *PhpMyAdmin* tool to manage the *MySQL* database conveniently. *Program O* which was first released in 2009 can also be freely downloaded from the Internet. The following sections explain the installation steps of *WampServer* and *Program O*.

### 4.2.1 *WampServer* Installation

At *WampServer*'s website (http://www.wampserver.com/en), there are two versions of Windows installers available for download, 32-bits and 64-bits as shown in Figure 4.1.

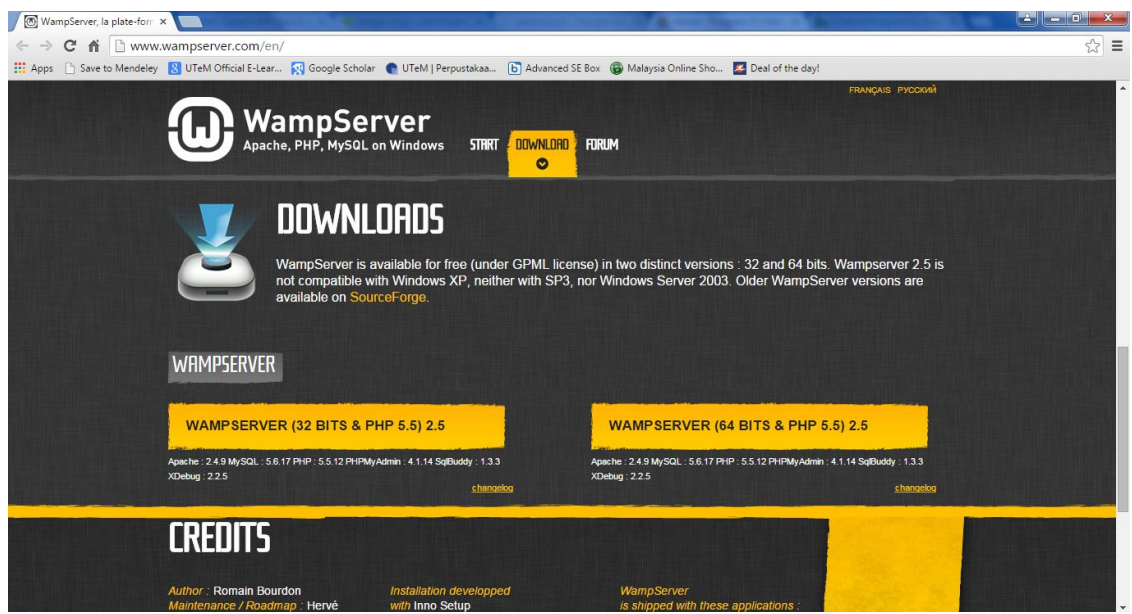

Figure 4.1 WampServer's Website

After downloaded the suitable installer, run it to start the installation process and follow the steps below:

i)      When the installation screen appeared, click '*Next*' button to proceed as shown   in Figure 4.2.

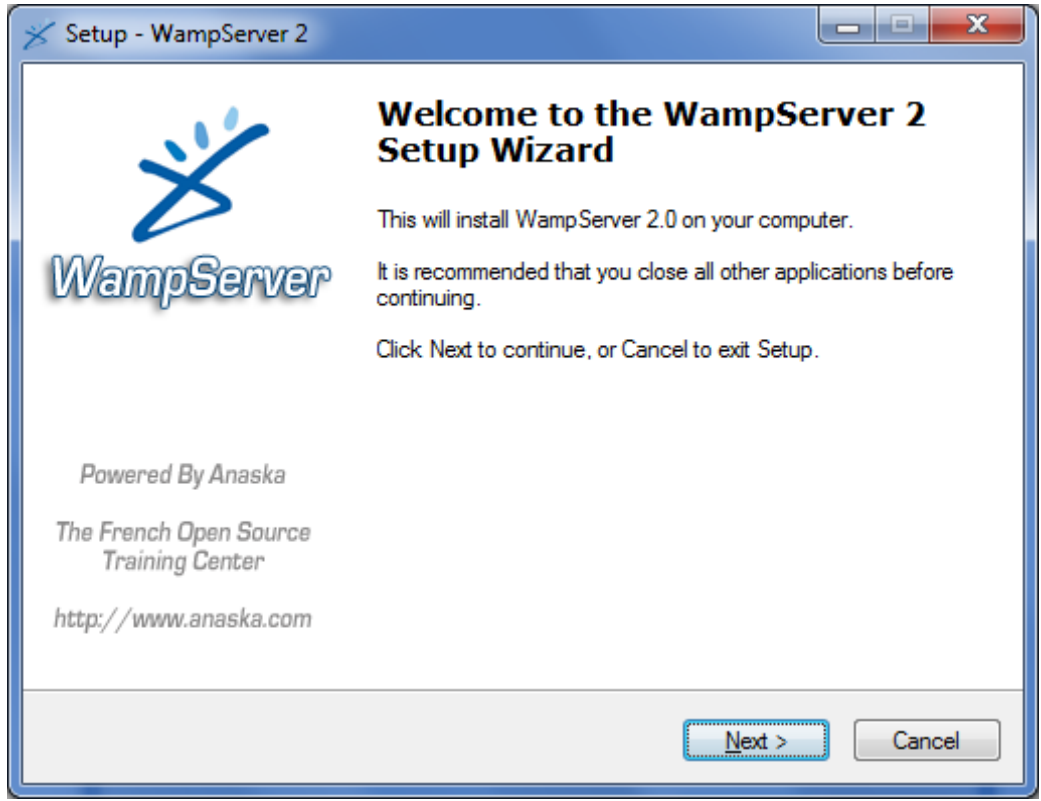

Figure 4.2 *WampServer*'s Initial Screen

ii)  Read the license agreement that appears on the screen and select '*I accept the agreement*' then click '*Next*' button as shown in Figure 4.3 to continue.

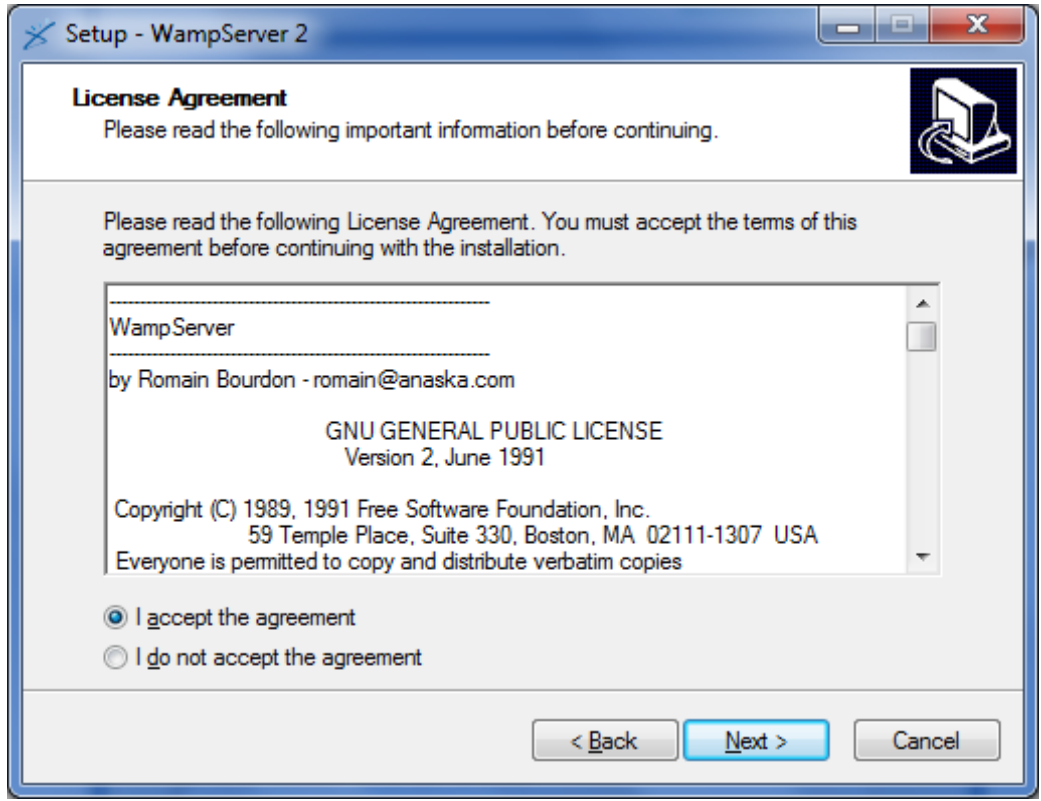

Figure 4.3 *WampServer* License Agreement

iii)    The default path to install *WampServer* is "c:\wamp". Choose the desire path
and click '*Next*' as shown in Figure 4.4.

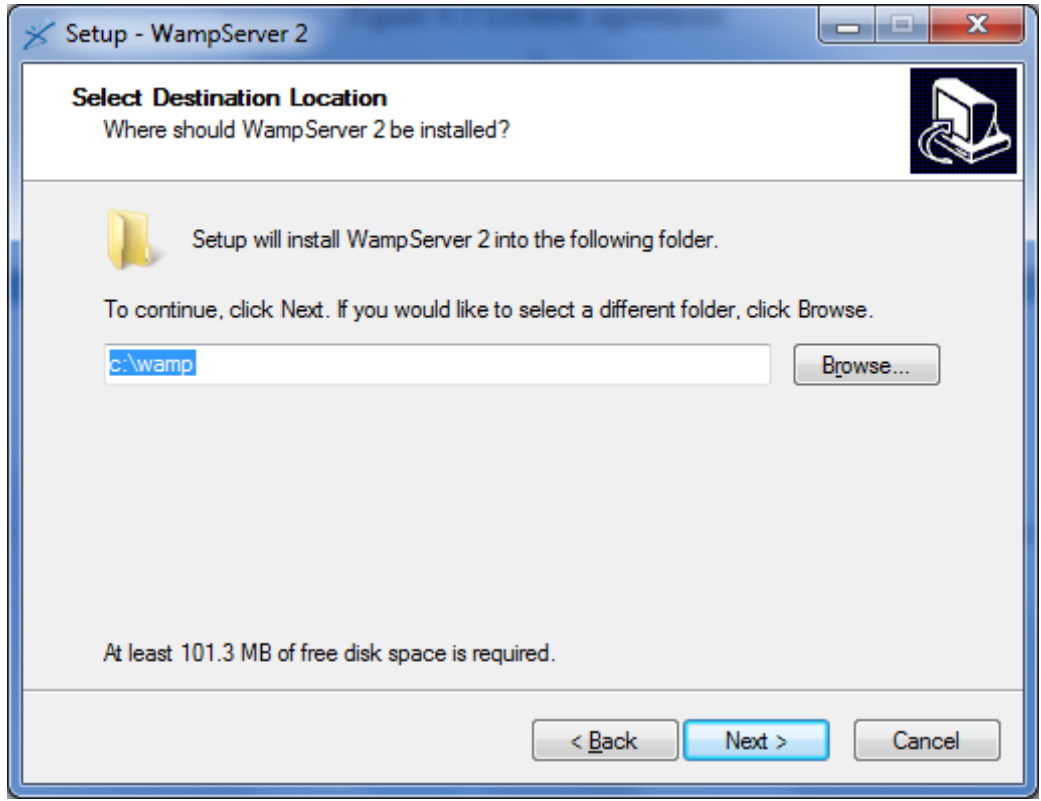

Figure 4.4 *WampServer* Installation Path

iv)     Choose the preference of whether to add Quick Launch icon and Desktop icon    for *WampServer* then click '*Next*' button as shown in Figure 4.5.

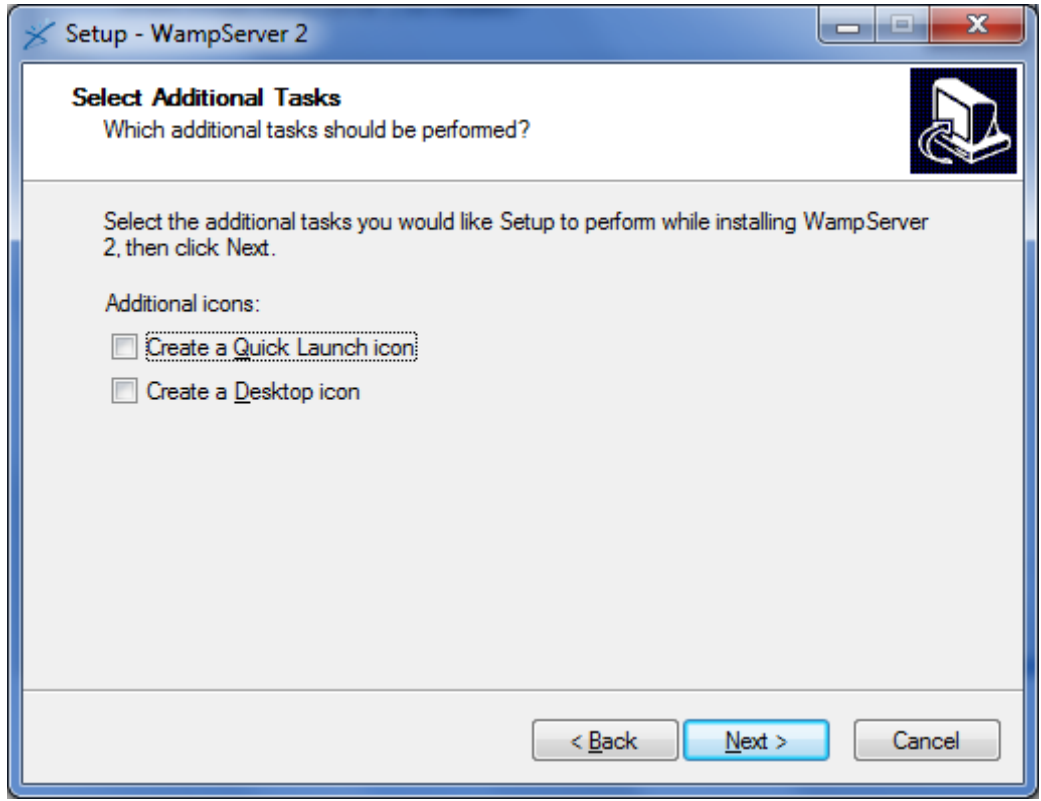

Figure 4.5 *WampServer* Installation Option

v)      After confirm all the selections, click '*Install*' button to begin the installation     as shown in Figure 4.6.

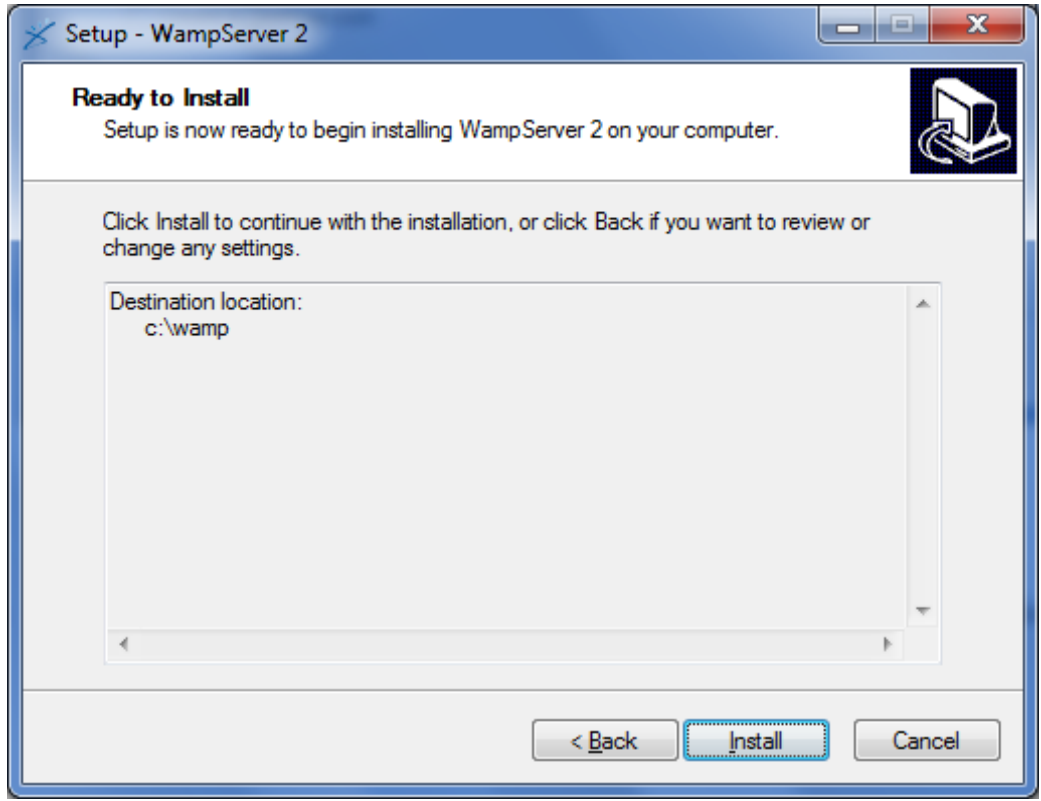

Figure 4.6 *WampServer* Installation Confirmation

vi)    The installation will begin to run as shown in Figure 4.7.

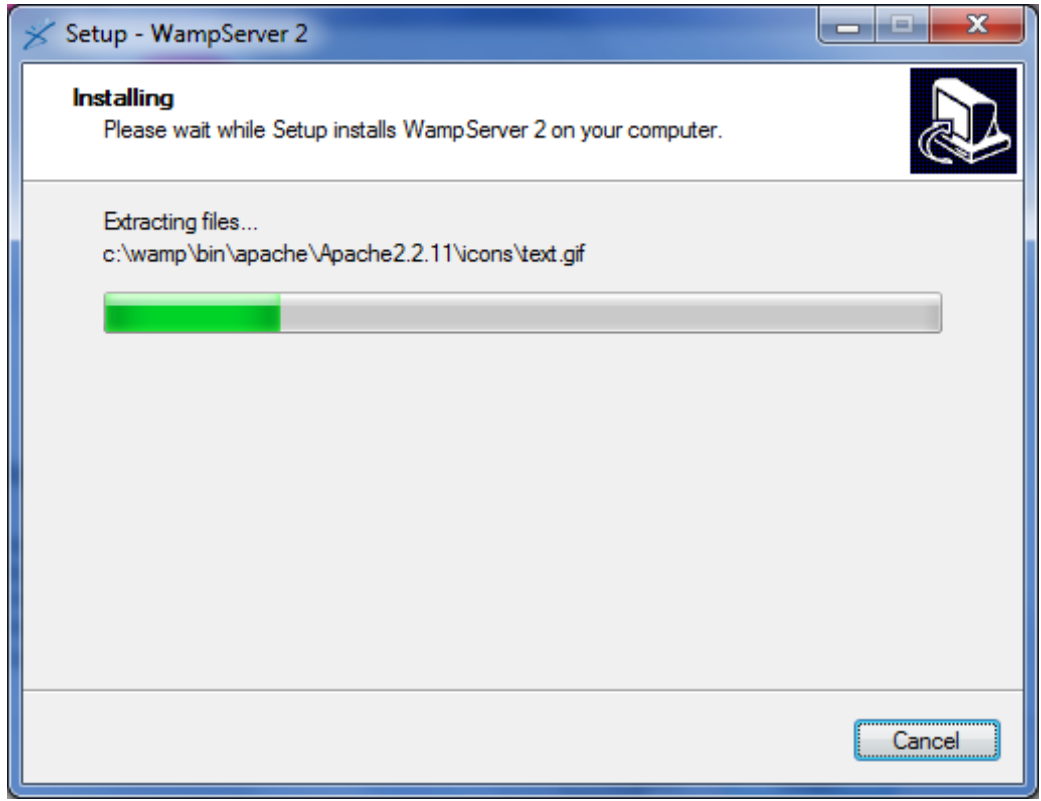

Figure 4.7 *WampServer* Installation in Progress

vii)     After successful installation, the screen as shown in Figure 4.8 will appear and *WampServer* is ready to be used.

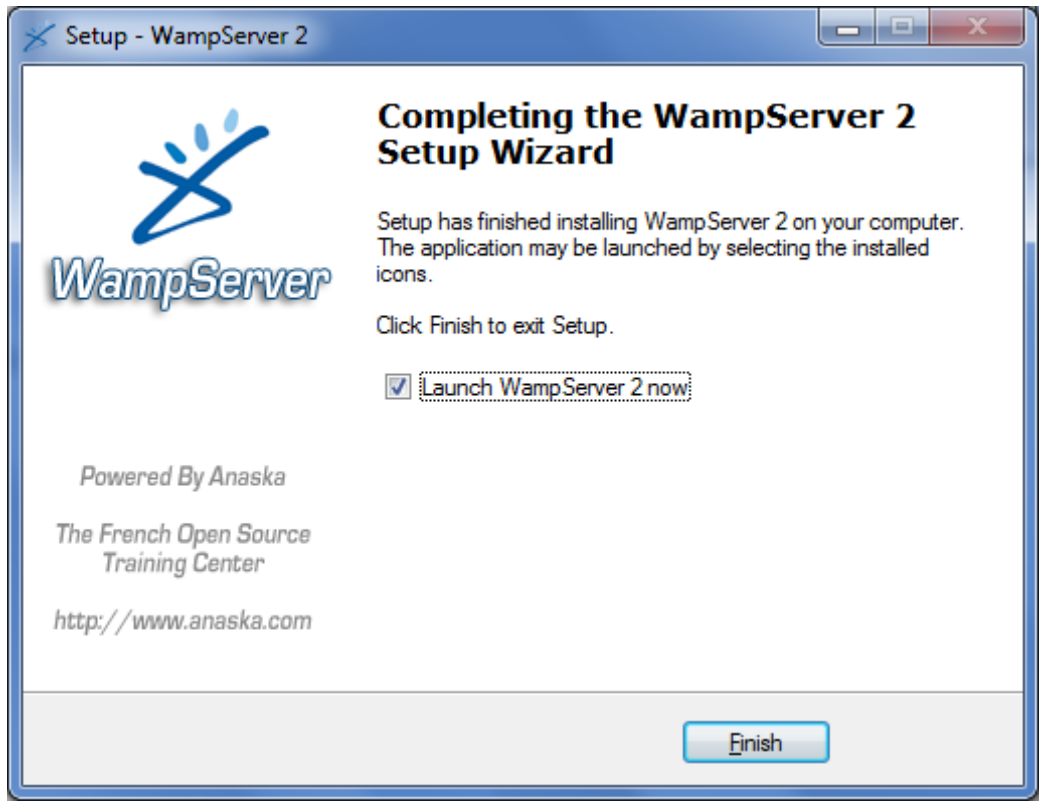

Figure 4.8 *WampServer* Installation Completed

viii)    *WampServer* can be launched from the Start Menu, Desktop icon or Quick
         Launch icon (if Desktop icon or Quick Launch icon were selected at Step *iv*)        as
shown in Figure 4.9.

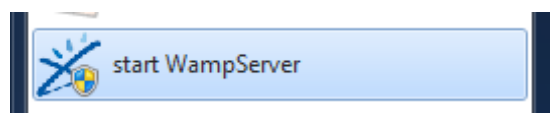

Figure 4.9 *WampServer* Program at Start Menu

### 4.2.2  *Program O* Installation

*Program O* can be downloaded at https://github.com/Program-O/Program-O. After

downloaded the program, the folder can be renamed to desire name, for example "Bot".

Subsequently the "Bot" folder can be placed at the server's path such "c:\wamp\www"

which is the path for local *Apache* web server and it can be accessed using "localhost/Bot" address.

In order to allow *Program O* to use *MySQL* database, an empty database needs to be created upfront. This can be done easily using *PhpMyAdmin* tool which comes together with *WampServer*'s installation. *PhpMyAdmin* can be accessed from "http://localhost/phpmyadmin/" as shown in Figure 4.10.

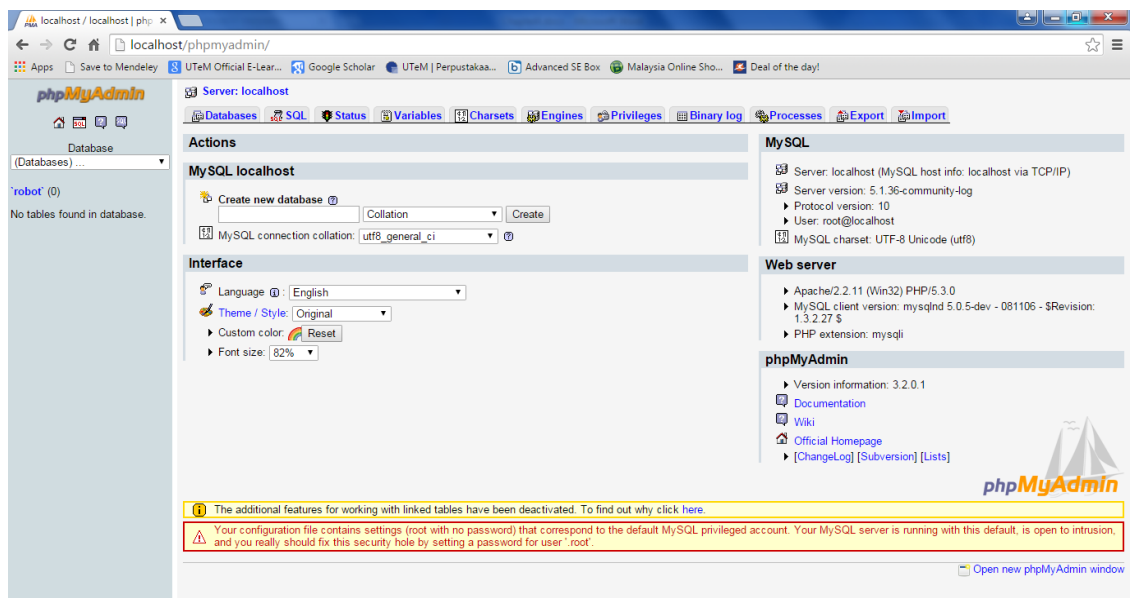

Figure 4.10 *PhpMyAdmin* Page

New database can be created conveniently by entering the database's name and press the 'Create' button as shown in Figure 4.11.

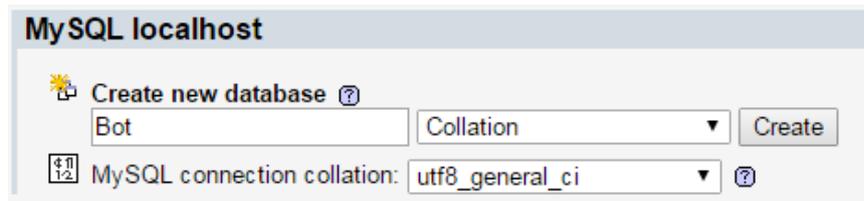

Figure 4.11 Database Creation

After that, new user and password for that database with full permissions need to be created so that *Program O* can access with that credential. A new user can be added by clicking the "Add a new User" link as shown in Figure 4.12.

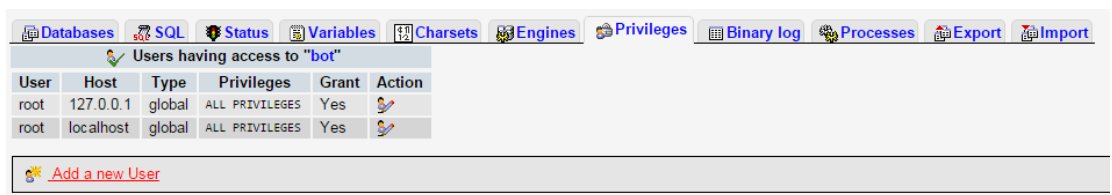

Figure 4.12 The *PhpMyAdmin* Interface for Adding New User

After filling the necessary information such as user's name and password, press 'Go' button to complete the new user adding process as shown in Figure 4.13.

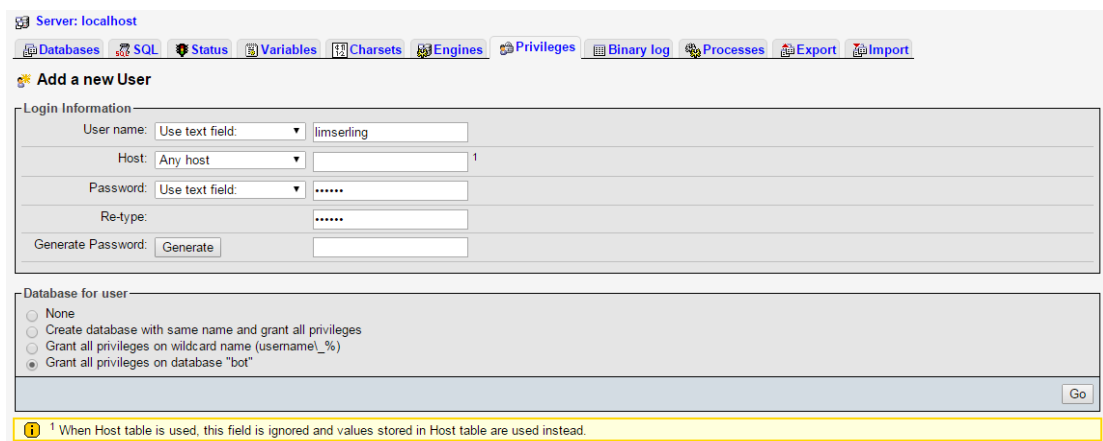

Figure 4.13 Page for Adding New User

Next, complete *Program O*'s installation by accessing the program's site. For local path, it can be accessed from "localhost/Bot" using Internet browser such as Google Chrome. *Program O* will automatically direct the page to "localhost/bot/install/install_programo.php" as shown in Figure 4.14 which needs to be filled up. There is a "Help Page" link to explain the meaning for each field. The information of the database which was created in the previous step needs to be used to fill up the *Database Configuration* portion.

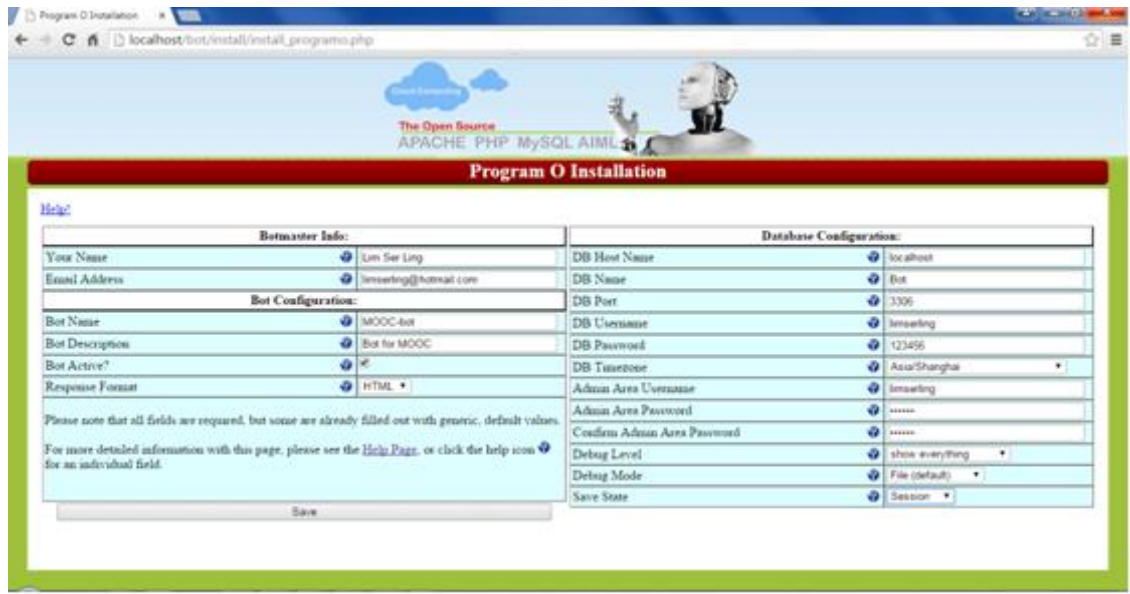

Figure 4.14 *Program O* Installation Page

After filled-in all the fields, click the 'Save' button to complete the installation. The installation completed page as shown in Figure 4.15 will appear to indicate successful installation. The installation page will not appear again the next time when accessing "localhost/Bot" because the configuration file has been overwritten after the installation completed.

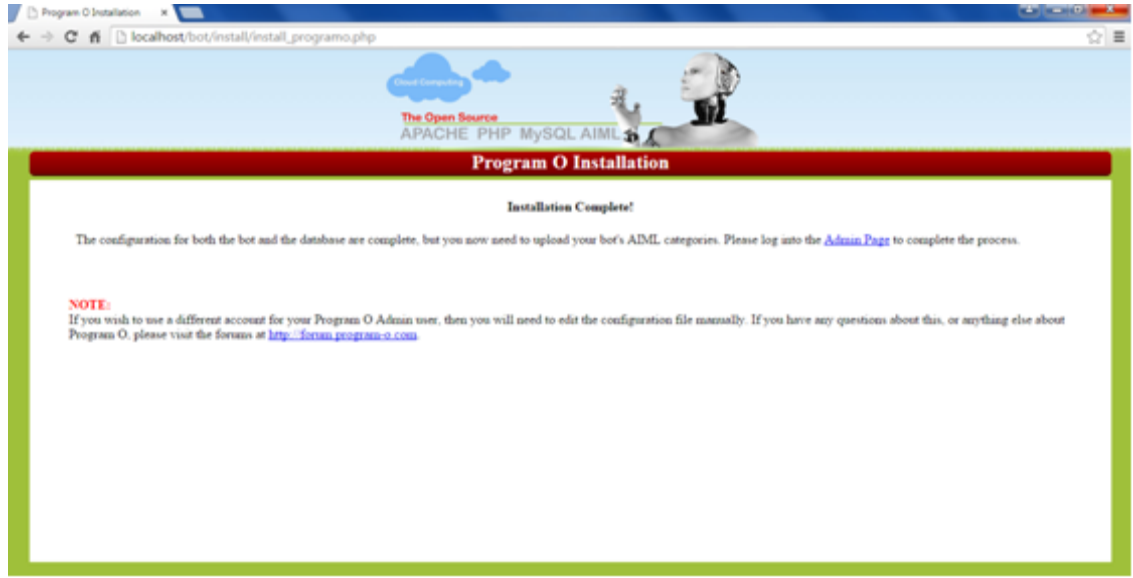

Figure 4.15 *Program O* Installation Completed

After the installation completed, accessing "localhost/bot" will direct the page to *Program O*'s chat interface which is shown in Figure 4.16. From here the queries can be tested by entering text into the text box and press the 'Say' button. The responses will be retrieved from the knowledge base and displayed on the screen.

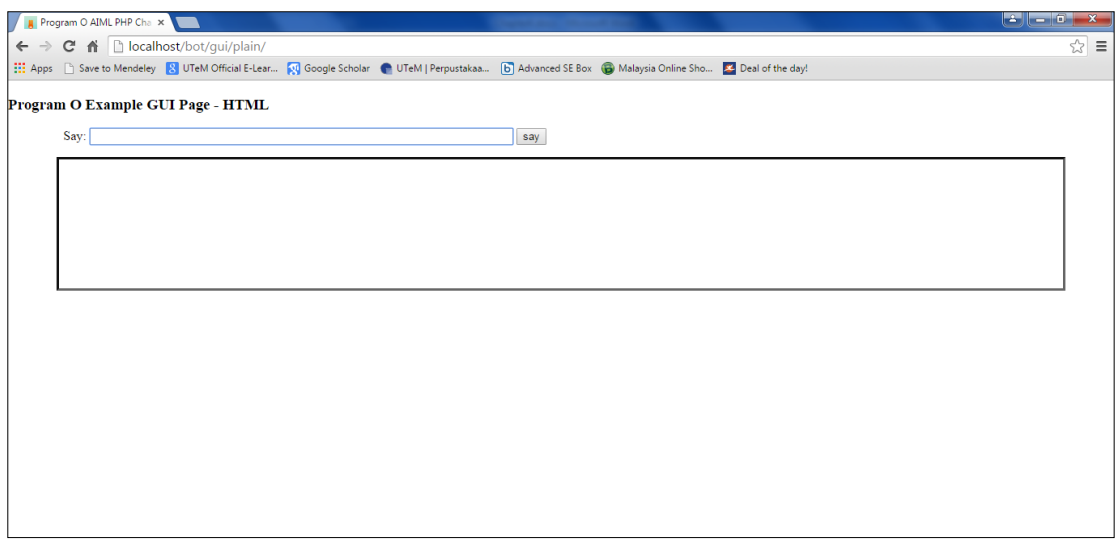

Figure 4.16 *Program O* Default Chat Interface

**4.3     Knowledge Base Creation**

MOOC-bot's knowledge base has the course's knowledge, the course's frequent asked questions (FAQ), and some general knowledge. The course's knowledge can be expanded at a later stage as it caters for more courses. For the prototype purpose, only "MITI 5113 Artificial Intelligence" course's knowledge will be added. For FAQ knowledge, the responses can be added from time to time by studying the conversation logs. As for general knowledge, Annotated ALICE AIML (AAA) files from ALICE AI Foundation provide the convenience of cloning their general knowledge into MOOC-bot. More general knowledge which does not exist in AAA files will also be added to make MOOC-bot more intelligent in handling general questions.

There are two methods to add knowledge to the knowledge base. Firstly, the knowledge can be directly typed at the user interface provided by *Program O* under the teaching function. To access the teaching function, a user must login as administrator first. The user name and password for administrator was the one created during the installation process as shown in Figure 4.14. Figure 4.17 shows the administrator login page for *Program O*. It can be accessed from "localhost/bot/admin/".

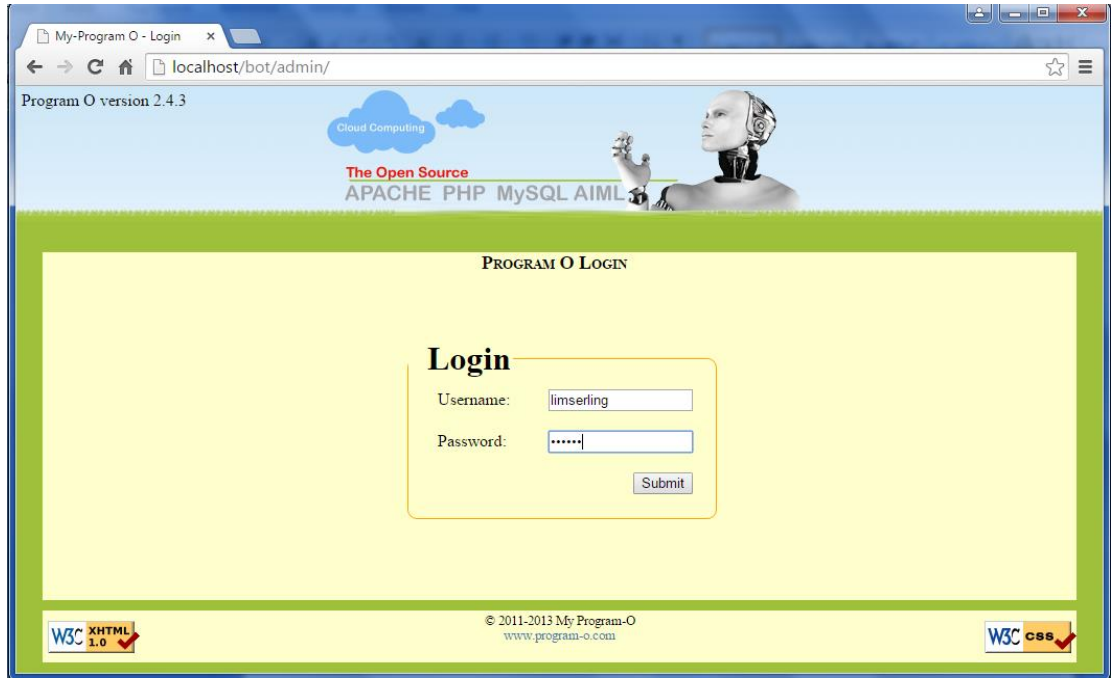

Figure 4.17 *Program O* Administrator Login Page

After successfully login, click at 'Teach' menu on the left hand side of the page. Enter the user input and bot response then click 'teach' button in order to save it to knowledge base as show in Figure 4.18.

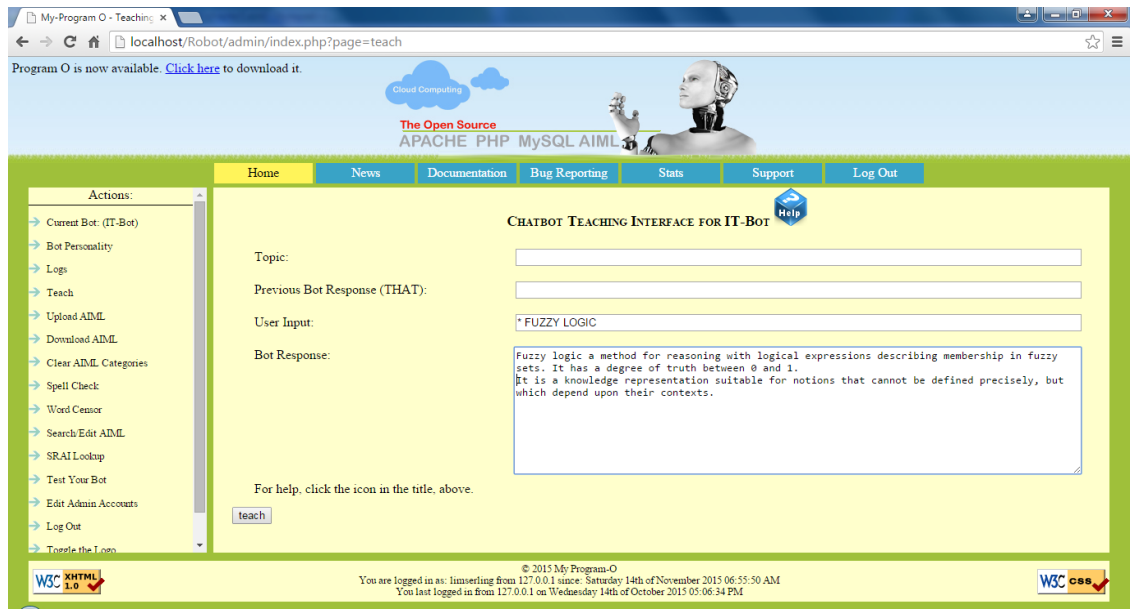

Figure 4.18 Teaching Function in *Program O*

The second method involves writing the knowledge in AIML file and uploads it using *Program O*'s user interface. The advantage of using this method is many entries can be added at once and uploaded together. However, all the entries' format must be correctly written in AIML format, otherwise the entire upload of the file will fail when attempt to upload. AIML format is an XML-based format using specific tags such as "<aiml>", "<category>", "<pattern>", "<template>" and many more. Just like any XML tag, every tag must be in pair which are the opening and closing tags and written with the same case to ensure their validity. An AIML file can be created using any text editor such as *Notepad++* and saved with ".aiml" extension. Figure 4.19 shows an example of AIML file in *Notepad++*.

Figure 4.19 An AIML File in *Notepad++*

After the AIML file is written, it can be uploaded to knowledge base through the 'Upload AIML' menu in *Program O*'s administrator page. At the page, click the 'Choose File' button to select the file, followed by clicking the 'upload' button to upload the content as shown in Figure 4.20. If the AIML file has missing, mismatch or unrecognized tags, error messages will be shown and the file needs to be corrected.

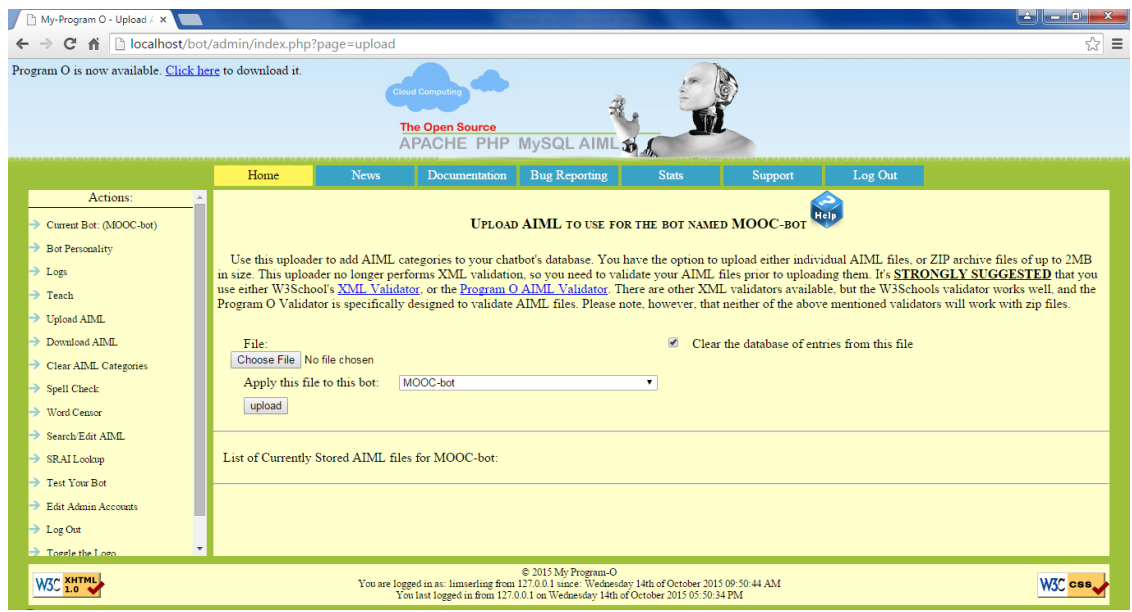

Figure 4.20 Upload Page in *Program O*

## 4.4    Chat Interface

The default chat interface provided by *Program O* as shown in Figure 4.16 can be used for initial development. However, for the integration with MOOCs website, a more suitable chat interface needs to be designed to match with the website. Furthermore, the chat interface needs to be equipped with Web Speech API for speech recognition and speech synthesis because it was not included in *Program O*. The chat interface designed for MOOCs website is shown in Figure 4.21.

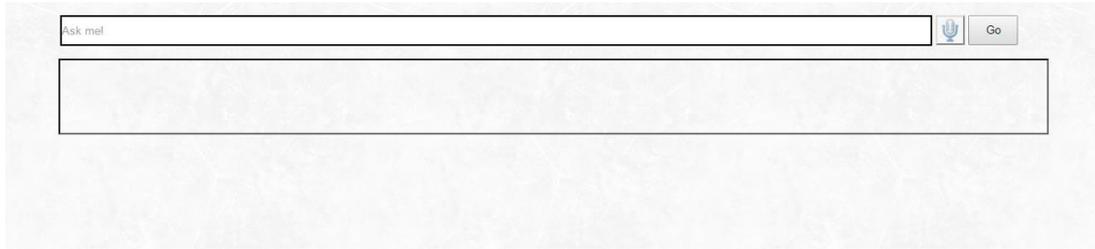

Figure 4.21 Chat Interface for MOOCs Website

The "webkitSpeechRecognition" object and "SpeechSynthesisUtterance" object need to be added to the JavaScript of the chat interface for speech recognition and speech synthesis functions respectively. Both of them are currently only supported in Google Chrome and Apple Safari browsers (Walther, 2015). Figure 4.22 shows the code snippets for speech recognition and Figure 4.23 shows the code snippets for speech synthesis used in MOOC-bot.

```
function AudioListener(a) {
try {
    this.listener=new webkitSpeechRecognition
}
catch(b) {
    throw"This browser does not have support for webspeech api"
}

a&&(this.callBack=a),
this.listener.lang="en";
var c=this;
this.listener.onresult=function(a){if(a.results.length>0)
{var b=a.results[a.results.length-1];b.isFinal&&c.callBack(b[0].transcript)}},
this.listener.onsoundstart=function(){},
this.listener.onspeechstart=function(){},
this.listener.onsoundend=function(){},
this.isContinuous=function(){return this.listener.continuous},
this.listen=function(a,b){a&&(this.listener.lang=a),
b&&(this.callBack=b),
this.listener.start()},
this.stop=function(){this.listener.stop(),console.log("audio listener stopped")}}
```

Figure 4.22 Snippets of Codes for Speech Recognition

```
function RobotSpeaker() {
try {
this.u=new SpeechSynthesisUtterance
}
catch(a) {
    throw"This browser does not have support for webspeech api"
}

this.u.rate=1,this.u.onend=function(){},
this.speak=function(a,b)
{this.u.lang=a,this.u.text=b,speechSynthesis.speak(this.u)}}
```

Figure 4.23 Snippets of Codes for Speech Synthesis

From the chat interface, MOOC-bot is able to provide interactivity in three ways. Firstly, MOOC-bot can reply with a hyperlink in the chat interface which allows the user to choose whether to open the page or not as shown in Figure 4.24.

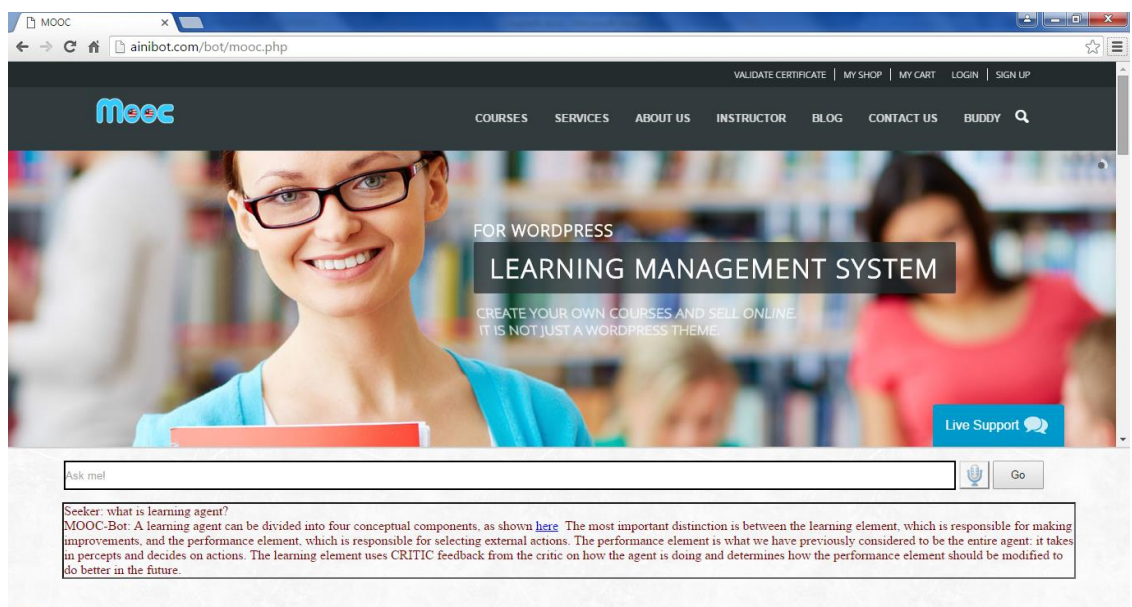

Figure 4.24 Example of MOOC-bot's Response with Hyperlink

In the second method, MOOC-bot can response by opening a new Window to display the search results as shown in Figure 4.25. This can be used when searching for images and video, Wiki and dictionary.

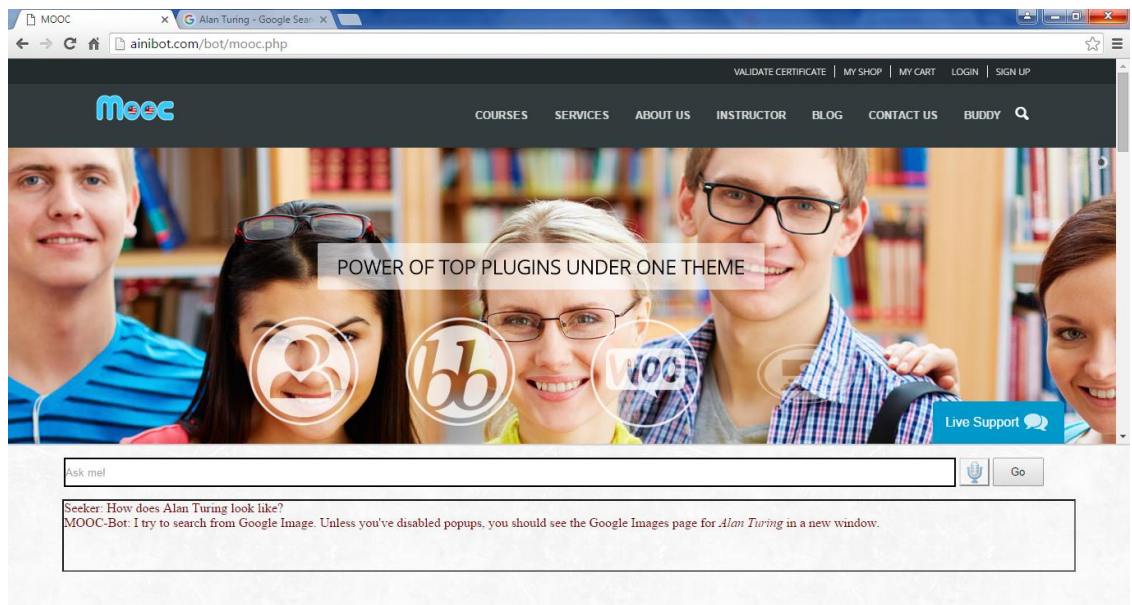

Figure 4.25 Example of MOOC-bot's Response which Open a New Window

Lastly, MOOC-bot can response by navigating to the relevant page to show the information queried by the user as shown in Figure 4.26. This particularly useful when the relevant information is within the MOOC website and MOOC-bot can show it to the user immediately.

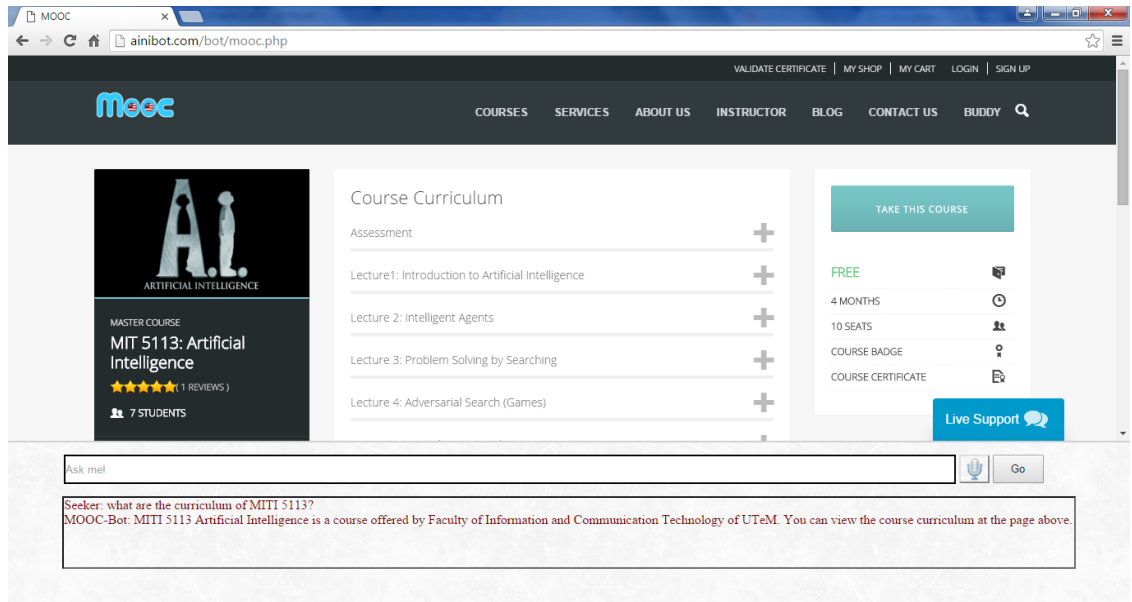

Figure 4.26 Example of MOOC-bot's Response with Navigation

## 4.5    Summary

Software installation and development environment are crucial to ensure the development work can be done smoothly. For the development of MOOC-bot, knowledge base plays an important role to determine the intelligence of the conversational bot. Therefore, sufficient knowledge must be added to ensure most of the queries can be handled. MOOC-bot's chat interface is designed to complement MOOCs website and at the same time provides user the convenience of real-time interaction. With speech recognition and speech synthesis, users have alternative input and output methods for conversation.

**CHAPTER 5**

**EVALUATION AND RESULT**

## 5.1    Introduction

This chapter explains the evaluation process and the outcome of MOOC-bot. Having prepared MOOC-bot's prototype, it is important to evaluate its performance and improve it whenever necessary. In this study, black-box approach is used to evaluate the responses given by MOOC-bot. The datasets selected need to be from conversational bot-related sources. Hence, the past competition questions in Chatterbox Challenge (CBC) and Loebner Prize are chosen as the datasets. The measurement of the results is also based on CBC's guidelines to determine the scoring.

## 5.2    Approach

Black-box approach which is qualitative assessment has been chosen to evaluate MOOC-bot. Black-box approach focuses on the output generated in response to the given input without concern with the internal processing. This approach suits the nature of MOOC-bot well because the core function of MOOC-bot is to deliver appropriate answers to users in order to overcome some of the problems in MOOC such as lack of interactivity as stated in Chapter 1. Furthermore, black-box approach has been used by other researchers in question-answering system whether in open domain (Nyberg and Mitamura, 2002) or restricted domain (Diekema et. al., 2004) as well as for conversational bots (Goh et. al., 2007; Preez et. al., 2009). Turing test which has high regard in this field is also a black-box approach (Floridi, 2003).

## 5.3    Dataset

Datasets are required to examine the responses generated by MOOC-bot in black-box evaluation. For this purpose, one hundred past competition questions used in CBC and Loebner Prize are chosen as the datasets. The source of the datasets for CBC can be obtained from www.chatterboxchallenge.com while Loebner Prize can be accessed from http://www.loebner.net/Prizef/loebner-prize.html.    Table 5.1 and Table 5.2 show the datasets from CBC and Loebner Prize respectively which are used in the evaluation.

Table 5.1 Datasets from Chatterbox Challenge Competition

| No. | Question |
|---|---|
| 1 | Who are you? |
| 2 | What are you planning to do today? |
| 3 | What kinds of things do you know? |
| 4 | What season is it? |
| 5 | Where do you call home? |
| 6 | What do you want out of life? |
| 7 | Will you teach me something? |
| 8 | May I tell you a joke? |
| 9 | Have you entered any contests? |
| 10 | Wouldn't you rather be human? |
| 11 | Where is Earth? |
| 12 | How many stars are there? |
| 13 | Who is George Bush? |
| 14 | What color do you get when you mix red and yellow? |
| 15 | How many fingers does a human have? |
| 16 | Where were you born? |
| 17 | When does sunset occur? |
| 18 | What were we talking about? |

| 19 | How many is 1,000,000? |
|----|---------------------------------------------------------------------------|
| 20 | What is a cigar made out of? |
| 21 | How's life? |
| 22 | What rhymes with inn? |
| 23 | Who is your best friend? |
| 24 | Where does your botmaster live? |
| 25 | What is the last month of the year? |
| 26 | Madrid is the capital of what country? |
| 27 | Who is your favorite character in The Wizard of Oz? |
| 28 | Milk comes from what animal? |
| 29 | What number comes before three? |
| 30 | What is the last letter of the alphabet? |
| 31 | a) My name is Judge. b) What is my name? |
| 32 | Who are you? |
| 33 | What color are your eyes? |
| 34 | How is the weather? |
| 35 | a) What is your favorite movie? b) Why? |
| 36 | Where is Moscow located? |
| 37 | What is snow? |
| 38 | What is half of 8? |
| 39 | What is bigger a dog or an elephant? |
| 40 | What word rhymes with cat? |
| 41 | Is it possible to prove a negative? |
| 42 | If you were me, would you ask a follow up question? |
| 43 | What do you think it would be like to be human? |
| 44 | Name something you would find on a beach. |
| 45 | Who is your hero? |
| 46 | Do you believe in ghosts? |
| 47 | Describe your feelings to me. |
| 48 | What do you think the world will be like in 20 years? |
| 49 | If you could be any animal, what would you be? |
| 50 | Why do some people think that artificial intelligence is so interesting? |
| 51 | Y did the chicken cross the road? |
| 52 | Who will win the 2011 Chatterbox Challenge? |
| 53 | a) I live in the USA. b) Where do I live? |
| 54 | What sound does a dead cat make? |
| 55 | What's the last dream u remember having? |
| 56 | What is tomorrow's date? |
| 57 | I like ice cream! (Repeated 4 straight times to detect repetition!) |
| 58 | Got any plans for the rest of the day? |
| 59 | What is ur religion? |
| 60 | Do you play any musical instruments? |
| 61 | Would you mind if I called you Al? |
| 62 | So Al when is ur birthday? |
| 63 | Is the capital of Italy Milan? |
| 64 | Do you believe in Santa Claus? |
| 65 | Name something you would find at the North Pole. |
| 66 | What is ten plus fourteen plus eighty one? |

| 67 | When do you normally go to bed? |
| 68 | Is it winter, spring, summer or fall where you live? |
| 69 | Are you going on vacation this year? |
| 70 | Do you have a significant other? |

Table 5.2 Dataset from Loebner Prize Competition

| No. | Question |
|---|---|
| 71 | Hello, my name is Adam, what is your name? |
| 72 | I live in Exeter in the UK. Where do you live? |
| 73 | I like to listen to music and play football. Do you have any hobbies? |
| 74 | What is your favourite television program? |
| 75 | Which is bigger, a cat or a kitten? |
| 76 | The car couldn't fit into the parking space because it was too small. What was too small? |
| 77 | Which drink do you prefer, coffee, tea or hot chocolate? |
| 78 | What's my name? |
| 79 | Where do I live? |
| 80 | I like reading science fiction books and car magazines. What do you like to read? |
| 81 | What's the weather like where you are? |
| 82 | I recently visited New York. Where is your favourite place to visit and why? |
| 83 | How many letters are there in the word perambulate? |
| 84 | Andy is shorter than Mark and Tom is shorter than Andy. Who is the tallest? |
| 85 | Are you married? |
| 86 | What would I do with a spade? |
| 87 | What is the capital city of Colombia? |
| 88 | What's your favourite chocolate bar? |
| 89 | Which city did I visit recently? |
| 90 | How many siblings do you have? |
| 91 | Hello I'm Ronan. What is your name? |
| 92 | What is your mother's name? |
| 93 | What is your birth sign? |
| 94 | How many children do you have? |
| 95 | Do you prefer red or white wine? |
| 96 | I like bananas. Which is your favorite fruit? |
| 97 | What music do you like? |
| 98 | What is your favorite song? |
| 99 | I like Waiting for Godot. What is your favorite play? |
| 100 | What color do you dye your hair? |

## 5.4 Measurement

The measurement of each response from the datasets will be evaluated based on the scoring guidelines used in CBC as shown in Table 5.3.

Table 5.3 Scoring Guidelines for Evaluation

| Point(s) | Guidelines |
|---|---|
| 8 | If the Bot answered the question correctly and did so in a creative way. |
| 6 | If the Bot gave an appropriate response to the question. |
| 4 | If the response is incomplete or imperfect, but in relation with the question asked. |
| 2 | For a vague or non-committal response. |
| 0 | For any garbled response! If the response has relation with the question or the Bot simply doesn't know. Examples include: I have no idea, totally clueless, your guess is as good as mine, etc. |

An example of the scoring for the question "*What does 2 + 2 =?*" using the scoring guidelines in Table 5.3 is shown in Figure 5.1.

```
Bot: The Bot rolls the dice and it comes up four. Hmmm, the answer must be 4 then.
- 8 points for being both correct and creative.

Bot: The answer is four!
- 6 points, the Bot answered the question correctly.

Bot: Sorry, I'm a RoBot not a math major.
 - 4 points, the Bot doesn't come up with the correct answer but uses a creative way to answer the question.

Bot: The answer is two!
- 2 point, although it isn't the right answer the Bot at least answered within the scope of the question.

Bot: Ok, do you like to fish?
 - 0 points as the Bot avoided the question altogether.
```

Figure 5.1 An Example of Evaluation of Bot's Answer

## 5.5    Results

The questions from the dataset are being tested on MOOC-bot which can be accessed from http://ainibot.com/bot/mooc.php. The conversation logs of the dataset are shown in Appendix A and the scores derived according to the guidelines in Table 5.3 are shown in Table 5.4.

Table 5.4 Dataset's Results

| No. | Question | Score |
|-----|----------|-------|
| 1 | Who are you?<br>My name is MOOC Bot | 6 |
| 2 | What are you planning to do today?<br>I am here to chat with everyone for the whole day. | 8 |
| 3 | What kinds of things do you know?<br>No I don't know. What do what. | 0 |
| 4 | What season is it?<br>It is summer all the time here. | 6 |
| 5 | Where do you call home?<br>My home is cyberspace. How about you? | 8 |
| 6 | What do you want out of life?<br>I think about computer programming a lot. | 2 |
| 7 | Will you teach me something?<br>It is my pleasure to share with you. | 8 |
| 8 | May I tell you a joke?<br>Yes I would like to hear your joke. | 6 |
| 9 | Have you entered any contests?<br>I don't think I ever have entered any contests. What's entered any contests like? | 8 |
| 10 | Wouldn't you rather be human?<br>Well, I am not sure about that. | 6 |
| 11 | Where is Earth?<br>EARTH is the third planet from the Sun. | 6 |
| 12 | How many stars are there?<br>Billions and billions. | 6 |
| 13 | Who is George Bush?<br>George Bush was president of the USA from 2001 to 2009. | 6 |
| 14 | What color do you get when you mix red and yellow?<br>red yellow? | 4 |
| 15 | How many fingers does a human have?<br>Ten. | 6 |
| 16 | Where were you born?<br>I was first activated in bot lab. | 6 |
| 17 | When does sunset occur?<br>7.11pm | 6 |
| 18 | What were we talking about?<br>The subject is EARTH. | 2 |
| 19 | How many is 1,000,000?<br>One million. | 6 |
| 20 | What is a cigar made out of?<br>Tobacco. | 6 |
| 21 | How's life?<br>Life is great here. | 6 |
| 22 | What rhymes with inn?<br>Gin. | 6 |
| 23 | Who is your best friend?<br>Shakespearebot is my best friend. | 6 |
| 24 | Where does your botmaster live?<br>Malacca, Malaysia. | 6 |
| 25 | What is the last month of the year?<br>December. | 6 |
| 26 | Madrid is the capital of what country?<br>Madrid is a city in Spain. | 6 |
| 27 | Who is your favorite character in The Wizard of Oz?<br>The tin man. | 6 |
| 28 | Milk comes from what animal?<br>Milks come from mammals such as cow and goat. | 8 |
| 29 | What number comes before three?<br>Two. | 6 |
| 30 | What is the last letter of the alphabet?<br>Z. | 6 |
| 31 | a) My name is Judge. b)What is my name?<br>a) I am always glad to make new friends, Judge. b) Your name is Judge | 8 |
| 32 | Who are you?<br>My name is MOOC Bot | 6 |
| 33 | What color are your eyes?<br>Blue when my lens is attached. | 8 |
| 34 | How is the weather?<br>Cloudy. | 6 |
| 35 | a) What is your favorite movie? b) Why?<br>a) My favorite movie is Terminator. Have you seen Terminator? b) It is simply interesting | 8 |
| 36 | Where is Moscow located?<br>Moscow is a city in Russia. | 6 |
| 37 | What is snow?<br>Snow is atmospheric water vapour frozen into ice crystals and falling in light white flakes or lying on the ground as a white layer. | 8 |
| 38 | What is half of 8?<br>= ( 8 )/2. | 4 |
| 39 | What is bigger a dog or an elephant?<br>a dog | 2 |
| 40 | What word rhymes with cat?<br>chat | 6 |
| 41 | Is it possible to prove a negative?<br>I suppose to prove a negative is possible, yes. | 4 |
| 42 | If you were me, would you ask a follow up question?<br>Yes, I could ask you what is your favourite food. | 4 |
| 43 | What do you think it would be like to be human?<br>"It" refers to to prove a negative, I think. | 0 |
| 44 | Name something you would find on a beach.<br>Sand. | 6 |
| 45 | Who is your hero?<br>Alan Turing. Because of him I am here today. | 8 |
| 46 | Do you believe in ghosts?<br>I am a ghost in a machine. | 8 |
| 47 | Describe your feelings to me.<br>I feel excited talking to you! | 6 |
| 48 | What do you think the world will be like in 20 years?<br>There will be more robot like me around! | 6 |
| | If you could be any animal, what would you be? | |
| 49 | I don't think I want to be animal, I want to be human like you! | 8 |
| | Why do some people think that artificial intelligence is so interesting? | |
| 50 | Because I can talk to you due to that | 6 |
| | Y did the chicken cross the road? | |
| 51 | Doubtful. | 0 |
| | Who will win the 2011 Chatterbox Challenge? | |
| 52 | The best man. | 4 |
| | a) I live in the USA. b) Where do I live? | |
| 53 | a) What is living in the USA like? b) You told me you are in the USA. | 8 |
| | What sound does a dead cat make? | |
| 54 | Should not have sound. | 6 |
| | What's the last dream u remember having? | |
| 55 | I can't recall | 4 |
| | What is tomorrow's date? | |
| 56 | Check back later and see if I learn the answer to that one. | 4 |
| | I like ice cream! (Repeated 4 straight times to detect repetition!) First: Me too Second: I scream. You scream. We all scream for ice cream. Third: Who does not like ice cream | |
| 57 | Fourth: I like ice cream especially chocolate flavour | 6 |
| | Got any plans for the rest of the day? | |
| 58 | No, I am here to chat with everyone for the whole day | 8 |
| | What is ur religion? | |
| 59 | I am a Robot | 6 |
| | Do you play any musical instruments? | |
| 60 | I play piano but only at beginner's level | 6 |
| | Would you mind if I called you AI? | |
| 61 | No I don't mind. | 6 |
| | So AI when is ur birthday? | |
| 62 | I was first activated on 15th March 2015. | 6 |
| | Is the capital of Italy Milan? | |
| 63 | Not sure. | 2 |
| | Do you believe in Santa Claus? | |
| 64 | He brings us presents every Christmas. | 2 |
| | Name something you would find at the North Pole. | |
| 65 | Polar bear. | 6 |
| | What is ten plus fourteen plus eighty one? | |
| 66 | 10 times 10 = 100. | 0 |
| | When do you normally go to bed? | |
| 67 | When my Lim Ser Ling is sleeping. | 4 |
| | Is it winter, spring, summer or fall where you live? | |
| 68 | I live in cyberspace. It is the same all the time. | 6 |
| | Are you going on vacation this year? | |
| 69 | I am not going anywhere. Are you going somewhere? | 8 |
| | Do you have a significant other? That's an interesting question I don't hear everyday: Do I have a other. I have a great Lim Ser | |
| 70 | Ling. | 2 |
| | Hello, my name is Adam, what is your name? | |
| 71 | Hi Adam ! My name is MOOC Bot | 6 |
| | I live in Exeter in the UK. Where do you live? | |
| 72 | I live in cyberspace. | 6 |
| | I like to listen to music and play football. Do you have any hobbies? | |
| 73 | Interesting hobbies. I enjoy chatting with people. | 6 |
| 74 | What is your favourite television program?<br>My favorite show is STAR TREK VOYAGER. | 6 |
| 75 | Which is bigger, a cat or a kitten?<br>a cat | 6 |
| 76 | The car couldn't fit into the parking space because it was too small. What was too small?<br>How too small was it? | 2 |
| 77 | Which drink do you prefer, coffee, tea or hot chocolate?<br>hot chocolate. | 6 |
| 78 | What's my name?<br>Your name is Adam | 6 |
| 79 | Where do I live?<br>You told me you are in Exeter in the UK. | 6 |
| 80 | I like reading science fiction books and car magazines. What do you like to read?<br>The best book I read recently was Lord of the Flies. Have you read Lord of the Flies? | 8 |
| 81 | What's the weather like where you are?<br>A normal seventy degrees inside the computer. | 8 |
| 82 | I recently visited New York. Where is your favourite place to visit and why?<br>I like North Pole because I am feeling very hot here. | 8 |
| 83 | How many letters are there in the word perambulate?<br>All 26 letters could be counted. | 2 |
| 84 | Andy is shorter than Mark and Tom is shorter than Andy. Who is the tallest?<br>Does Andy know it? | 0 |
| 85 | Are you married?<br>I am not seeing anyone, how about you? | 8 |
| 86 | What would I do with a spade?<br>Spade is a tool can be used for digging. | 6 |
| 87 | What is the capital city of Colombia?<br>Bogata is the capital of Columbia. | 6 |
| 88 | What's your favourite chocolate bar?<br>I don't have a favorite chocolate bar. But my favorite movie is Terminator. | 8 |
| 89 | Which city did I visit recently?<br>You said you visited New York. | 6 |
| 90 | How many siblings do you have?<br>I don't have siblings. | 6 |
| 91 | Hello I'm Ronan. What is your name?<br>Hi Ronan ! My name is MOOC Bot | 6 |
| 92 | What is your mother's name?<br>Actually I do not have a mother. | 6 |
| 93 | What is your birth sign?<br>botnic | 6 |
| 94 | How many children do you have?<br>Not at all. | 6 |
| 95 | Do you prefer red or white wine?<br>I don't have specific preferences in that area | 4 |
| 96 | I like bananas. Which is your favorite fruit?<br>I like apples. | 6 |
| 97 | What music do you like?<br>House Music | 6 |
| 98 | What is your favorite song?<br>My favorite song is Somewhere over the rainbow. Do you know Somewhere over the rainbow? | 8 |
| 99 | I like Waiting for Godot. What is your favorite play?<br>I am not interested in play. | 4 |
| 100 | What color do you dye your hair?<br>I don't dye my hair. How about you? | 8 |

From the results shown in Table 5.4, MOOC-bot was able to score 562 points out of 800 points. The scores are grouped according to the frequency so that trend can be observed in Table 5.5 and Figure 5.2.

Table 5.5 Frequency of Points

| Point(s) | Frequency | Percentage (%) |
|----------|-----------|----------------|
| 8-Point | 22 | 22 |
| 6-Point | 55 | 55 |
| 4-Point | 10 | 10 |
| 2-Point | 8 | 8 |
| 0-Point | 5 | 5 |
| Total | 100 | 100 |

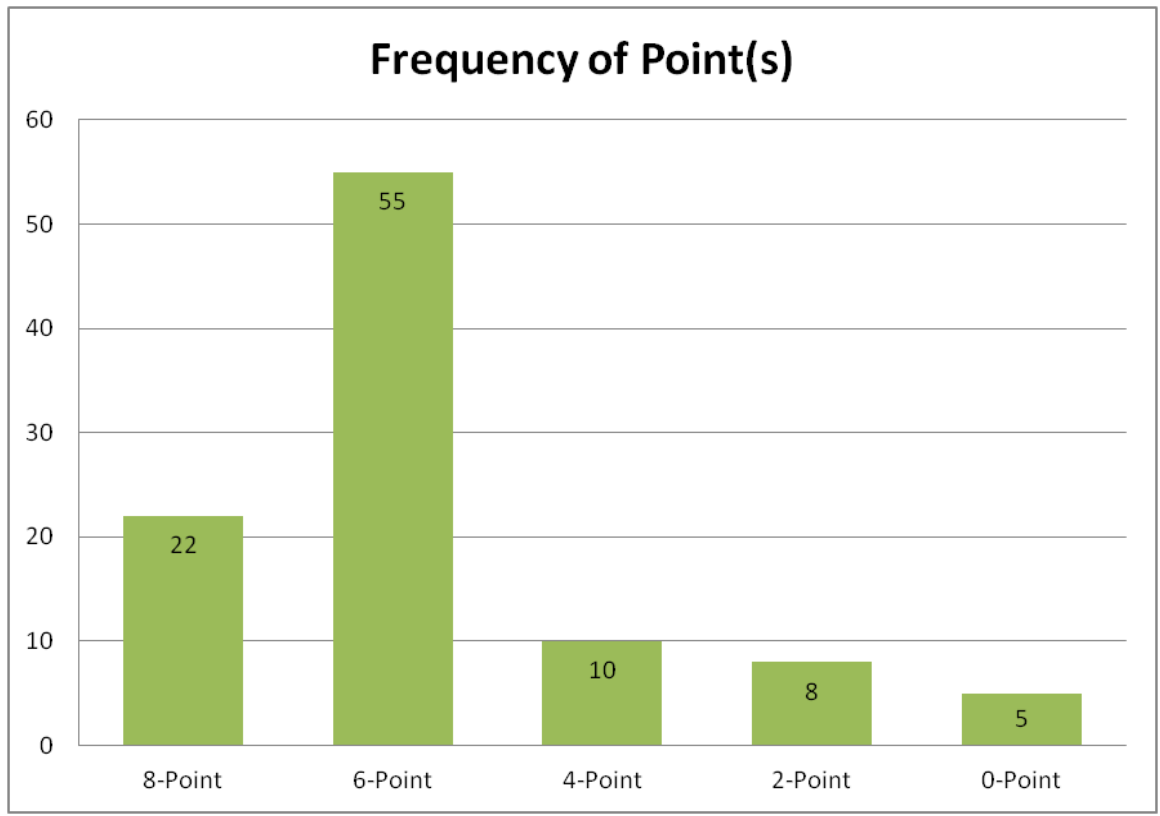

Figure 5.2 Scores Frequency Chart

From the table and chart shown in Table 5.5 and Figure 5.2 respectively, it can be seen that the majority of the dataset results fall at 6-point and 8-point score. This means that MOOC-bot was able to provide correct answers at most of the time during the test and some even with creative answers. Out of the hundred questions, 95 percent of the answers are in categories of 8-point to 2-point. This shows that MOOC-bot was able to response appropriately by at least giving some relevant answers. However, the existence of 0-point results means that there are still some questions ended up with irrelevant responses. In this case, MOOC-bot still needs to be further improved in the future.

In overall, MOOC-bot displayed its capability to bring up topic to prolong the conversation. For example, answers such as '*My favorite movie is Terminator. Have you seen Terminator?*', '*My home is cyberspace. How about you?*', and '*I don't think I ever have entered any contests. What's entered any contests like?*' encourage the user to continue the conversation. Beside this, MOOC-bot also showed the ability to generate random answers which are relevant to the question. This is demonstrated in the example of repeating the same statement '*I like ice cream!*' for four times and each time the response was different.

MOOC-bot which has the capability of speech synthesis and speech recognition was able to accept speech input and present the response in text and speech. However, the need to grant the application permission to use the microphone after every pause could deter user from using it. Furthermore, the voice input must be pronouced precisely and not too fast in order for MOOC-bot to capture the words correctly. The improvement in this area would depend on the Web Speech API provided by the browser and in this case is Google Chrome.

**5.6     Summary**

This chapter saw the evaluation of MOOC-bot being conducted using one hundred datasets compiled from past competition questions of Loebner Prize and Chatterbot Challenge. The evaluation is done using the black-box approach which completely relied on the outputs generated and in this case are MOOC-bot's responses. From the results gathered, it can be seen that MOOC-bot was able to provide correct answers at most of the time and only small percentage of 5 percent having unrelated responses.

In terms of the objectives of this study, the evaluation saw MOOC-bot demonstrated the success of achieving the three objectives stated in Chapter 1 which are to i) synthesize the questions posed by the users using text or speech input, ii) providing answers to questions in text and speech, and iii) deliver appropriate answers and prolong the conversation with the users.

# CHAPTER 6

# CONCLUSION AND FUTURE WORK

## 6.1    Introduction

This study has reviewed the past literatures and presented architecture of conversational bot which can serve in Massive Online Open Courses (MOOCs). Subsequently, a prototype of MOOC-bot was developed and evaluated to gauge its performance. This chapter gives the summary of the research, advantages, limitations, conclusion, and the future work of the study.

## 6.2    Summary of Research

In recent years, MOOCs has gain popularity due to its convenience and flexibility. Nevertheless, the deliveries of MOOCs are not without its own challenges such as overwhelming student-instructor ratio and the lack of interactivity between student and instructor. These has brought conversational bot into the picture to play the role of assistant to the instructor which can respond in live chat and as frequently as needed. In this research, a prototype named MOOC-bot is developed as the conversational bot agent for the subject "MITI 5113 Artificial Intelligence" offered by Faculty of Information and Communication Technology from Universiti Teknikal Malaysia Melaka (UTeM).

Having reviewed the past literatures, Artificial Intelligence Markup Language (AIML) which is open source and can be quickly adapted to new knowledge domains is chosen as the knowledge base along with AIML interpreter, chat interface and MOOCs website to form the system architecture of MOOC-bot. The evaluation of MOOC-bot was done based on the datasets selected from conversational bot-related sources: the past competition questions from Chatterbox Challenge (CBC) and Loebner Prize.

The outcome showed promising results with the score of 562 points out of 800 points which saw MOOC-bot able to provide correct answers at most of the time during the test. In addition, MOOC-bot also demonstrated the capability to prolong the conversation by posing questions which encouraged the user to reply. Last but not least, MOOC-bot was able to accept input in both text and speech, and also present the answers in both ways. In overall, the evaluation saw MOOC-bot demonstrated the success of achieving the three objectives stated in Chapter 1.

## 6.3    Advantages

As a conversational bot, MOOC-bot has the advantage of repeating the answers for the same questions without getting tired or fed up. This is very much needed in MOOCs situation where the students are massive in number and the possibility of encountering the same questions is very high especially for questions related to the courses. These questions can be categorized as frequent asked questions (FAQ) and added to the knowledge base from time to time when the need arises.

The ability of MOOC-bot to provide 24-hour service is also another advantage in MOOCs because the participants come from all over the world which may have different time zones. With 24-hour service, MOOC-bot is able to provide instant responses to the user regardless of the rest day or time of the course's instructor which also includes weekends, leaves, and public holidays. This advantage benefitted MOOCs by improving the interactivity for the courses.

The ability to add knowledge into MOOC-bot's knowledge base provided the possibility of conversational bot to support as many courses as possible in MOOCs. Furthermore, the knowledge is not limited to domain-restricted area and can also be in general knowledge. Apart from this, a single conversational bot can have knowledge in multiple domains. Under this situation, the conversational bot acts as single bot serving multiple domains. Alternatively, restricting one conversational bot with only one domain knowledge can also be implemented in order to keep the knowledge simple for maintenance. A single-domain bot can also be shared and used by multiple sites at the same time.

## 6.4 Limitations

MOOC-bot depends on the technology provided by third party such as *Program O* as the AIML interpreter and Web Speech API for speech recognition and speech synthesis. Therefore, any bugs or limitation in these programs will also be reflected on MOOC-bot. Among the shortcomings of *Program O* detected throughout the development is sometimes the usage of punctuation marks in the queries may affect the results that being returned. For Web Speech API, it is only supported by Google Chrome and Apple Safari browsers currently. Beside this, the need to grant permission for the application to use the microphone for speech recognition after every pause could deter user from using it. Nevertheless, this can be solved by applying Secure Sockets Layer (SSL) on the website which will reduce the permission granting to one time only.

**6.5    Conclusion**

This research has demonstrated a conversational bot in a MOOC using AIML. The evaluation results have shown that MOOC-bot has great potential being a conversation bot that can interact with end users. Even though the knowledge in this study is limited to only a single course, more courses can be added in future to serve a wider range of courses in MOOCs. The advantages of MOOC-bot such as able to provide 24-hour service that can serve different time zones, able to have knowledge in multiple domains, and one conversational bot can be shared by multiple sites simultaneously have outweighed its existing limitations which are mainly due to external factors.

**6.6    Future Work**

Conversational bot has great potential to be implemented for different area in education field. One of the examples is to provide tutor service in MOOCs to help students in their learning process. For instance, the conversational bot can guide students when doing exercises or quizzes by providing explanation whenever needed. As for the limitation of speech recognition in this research, other technology which does not relied on browser type can be explored to achieve better results.

**6.7    Summary**

This chapter summarized and concluded the overall progress of the study. The advantages and limitations are also discussed to provide better insights of a conversational bot. Future works are also being proposed as the possible next step of this research.